\documentclass{article}
\usepackage{ijcai21}

\RequirePackage[T1]{fontenc} 
\RequirePackage[tt=false, type1=true]{libertine} 
\RequirePackage[varqu]{zi4} 

\RequirePackage[libertine]{newtxmath}
\usepackage[letterspace=-25]{microtype}

\usepackage{soul}
\usepackage{url}
\usepackage[hidelinks]{hyperref}
\usepackage[utf8]{inputenc}
\usepackage[small]{caption}
\usepackage{graphicx}
\usepackage{amsmath}
\usepackage{booktabs}
\usepackage{savesym}
\savesymbol{Bbbk}
\savesymbol{openbox}

\usepackage{amsmath}
\usepackage{amsthm}
\usepackage{amsfonts}
\usepackage{hyperref}
\usepackage{float}
\usepackage{graphicx}
\usepackage{tikz}
\usepackage{stackengine}

\usepackage{amssymb}
\usepackage{amsthm}
\usepackage{MnSymbol}
\usepackage{color}
\usepackage{url}
\usepackage{multirow}
\usepackage{xcolor}
\usepackage{dsfont}
\usepackage{algorithm}
\usepackage[noend]{algpseudocode}

\restoresymbol{NEW}{Bbbk}
\restoresymbol{NEW}{openbox}

\usepackage[switch]{lineno}

\usepackage{tablefootnote}

\usepackage{subcaption}

\usepackage{amsmath}
\usepackage{amsthm}
\usepackage{amsfonts}
\usepackage{hyperref}
\usepackage{float}
\usepackage{graphicx}
\usepackage{tikz}
\usepackage{stackengine}

\newcommand{\model}{\mathcal{M}}
\newcommand{\network}{\mathcal{M}}
\newcommand{\intervalnet}{\mathcal{I}}

\newcommand{\abst}[2]{\mathcal{I}_{(#1,#2)}}

\newcommand{\card}[1]{\lvert#1\rvert}

\renewcommand{\powerset}[1]{\mathbb{I}(#1)}

\DeclareMathOperator{\argmax}{arg\,max}
\DeclareMathOperator*{\argmin}{arg\,min}

\newcommand{\mshift}{S}
\newcommand{\distance}[3]{\lVert #1 - #2\rVert_{#3}}
\newcommand{\delequal}{\mathrel{\ensurestackMath{\stackon[1pt]{=}{\scriptscriptstyle\Delta}}}}

\newcommand{\dataset}{\mathcal{D}}

\newcommand{\name}{\Delta}

\newcommand{\AR}[1]{\textcolor{black}{#1}}
\newcommand{\FL}[1]{\textcolor{black}{#1}}
\newcommand{\FT}[1]{\textcolor{black}{#1}}
\newcommand{\JJ}[1]{\textcolor{black}{#1}}

\newtheorem{definition}{Definition}

\newtheorem{lemma}{Lemma}
\newtheorem{remark}{Remark}
\newtheorem{example}{Example}

\algnewcommand\algorithmicforeach{\textbf{for each}}
\algdef{S}[FOR]{ForEach}[1]{\algorithmicforeach\ #1\ \algorithmicdo}

\usepackage{fancyhdr}
\fancypagestyle{firstpage}{%
    \fancyhf{}
    \setlength{\headsep}{0.3in}
    \cfoot{\thepage}
}
\fancypagestyle{otherpages}{%
    \fancyhf{}
    \setlength{\headsep}{0.3in}
    \rhead{Paper Title}
    \lhead{Authors}
    \cfoot{\thepage}
}

\title{Formalising the Robustness of Counterfactual Explanations for \\Neural Networks}

\renewcommand*{\thefootnote}{\fnsymbol{footnote}}
\author {
    Junqi Jiang\footnote{These authors contributed equally.},
    Francesco Leofante\footnotemark[1],
    Antonio Rago,
    Francesca Toni\\
\affiliations\large
    Department of Computing, Imperial College London, UK\\
\emails
    \{junqi.jiang20, f.leofante, a.rago15, f.toni\}@imperial.ac.uk
}
\begin{document}

\renewcommand*{\thefootnote}{\arabic{footnote}}
\setcounter{footnote}{0}

\maketitle

\begin{abstract}
The use of counterfactual explanations (CFXs) is an increasingly popular explanation strategy for machine learning models. However, recent studies have shown that these explanations may not be robust to changes in the underlying model (e.g., following retraining), which raises questions about their reliability in real-world applications. Existing attempts towards solving this problem are heuristic, and the robustness to model changes of the resulting CFXs is evaluated with only a small number of retrained models, failing to provide exhaustive guarantees. {To remedy this}, we propose \emph{$\name$-robustness}, the first notion to formally and deterministically assess the robustness (to model changes) of CFXs for neural networks. We introduce an abstraction framework based on interval neural networks 
to verify the $\name$-robustness of CFXs against a possibly infinite set of changes to the model parameters, i.e., weights and biases. We then demonstrate the utility of this approach in two distinct ways. First, we  analyse the $\name$-robustness of a number of CFX generation methods from the literature and show that they unanimously host significant deficiencies in this regard. Second, we demonstrate how embedding $\name$-robustness within existing methods can provide CFXs which are provably robust. 
\end{abstract}

\section{Introduction}
\label{sec:intro}

Ensuring that machine learning models are explainable has become a dominant goal in recent years, giving rise to the field of \emph{explainable AI} (XAI). One of the most popular strategies for XAI is the use of \emph{counterfactual explanations} (CFXs) (see~\cite{Stepin_21} for an overview), favoured for a number of reasons including their intelligibility~\cite{Byrne_19}, appeal to users~\cite{Barocas_20}, information capacity~\cite{Kenny_21} and alignment with human reasoning~\cite{Miller_19}. 
A CFX for a given input to a model is defined as an altered input 
\FT{for which the model} gives a different output to that of the original input.
Consider the classic illustration of a loan application, with features \emph{unemployed} 
status, \emph{25} years of age and \emph{low} credit rating, being classified by a bank's 
model as rejected. A CFX for the rejection could be an altered input where a \emph{medium} credit rating (with the other features unchanged) would result in the loan being accepted, thus giving the applicant an idea of what is required to change the output.
Such correctness of the 
modified output in attaining an alternative value is the basic property of CFXs, referred to as \emph{validity}, and is one of a whole host of metrics around which CFXs are designed \FT{(e.g., see \cite{guidotti2022counterfactual})}.


Our main focus in this paper is the metric of \emph{robustness}. This is
most often defined as \emph{robustness to input perturbations}, i.e.,  the validity of CFXs when perturbations are applied to inputs~\cite{Sharma_20}.
While this notion is useful, e.g., for protecting against manipulation~\cite{Slack_21}, other forms of robustness can be equally important in ensuring that CFXs are safe and can be trusted.
\emph{Robustness to model changes}, i.e., the validity of CFXs when model parameters are altered, has thus far received little attention but is arguably one of the most commonly required forms of robustness, given that model parameters change every time retraining occurs~\cite{Rawal_20X}.
Indeed, if a CFX is invalidated with just a slight change of the training settings as in, e.g.,~\cite{Dutta_22}, we may question its quality in terms of real-world meaning.
Consider the loan example: if, after retraining, the loan applicant changing their credit rating to \emph{medium} no longer changes the output to accepted (thus invalidating the CFX), the CFX was not robust to the model changes induced during retraining. In this case, it might be argued that the bank should have a policy to guarantee that this CFX remains valid regardless, but this may have unfavourable consequences for the bank. Therefore, it is desirable that the CFXs account for such robustness.

Though some have targeted robustness to model changes\footnote{Referred to simply as \emph{robustness}, unless otherwise specified
.}, e.g., ~\cite{UpadhyayJL21,Dutta_22}, these approaches are heuristic, and may fail to provide strong robustness guarantees. 
Formal methods for assessing CFXs along this metric are lacking. Indeed, there are calls for both formal explanations for non-linear models such as neural networks~\cite{Ignatiev_22} and for standardised benchmarking in evaluating CFXs~\cite{Kenny_21}, voids we help to fill.

In this work we propose the novel notion of \emph{$\name$-robustness} for assessing the robustness of CFXs for neural networks in a formal, deterministic manner. 
We introduce an abstraction framework based on \emph{interval neural networks}~\cite{PrabhakarA19} to verify the robustness of CFXs against a possibly infinite set of changes to the model parameters, i.e., weights and biases. 
This abstraction allows for a set of parameterisable shifts, $\name$, in the model parameters, permitting users to tailor the strictness of robustness (depending on the application). For illustration, consider the loan example once more: the bank knows the scale of typical changes in their models and could encode this into $\Delta$. The bank would then be able to provide only $\name$-robust CFXs such that they are valid under any expected model shift during retraining (and if 
a model shift exceeds $\name$, they would have been alerted to this fact). 
It can be seen, even
from this simple example, that $\name$-robustness can provide priceless
guarantees in high-stakes or sensitive situations.

After covering related work (§\ref{sec:related}) and the necessary preliminaries (§\ref{sec:prel}), we make the following contributions.

\begin{itemize}
    \item We introduce a novel notion of $\name$-robustness of CFXs for neural networks and propose an abstraction framework based on interval neural networks to reason about it (§\ref{sec:robust_cfx}).
    \item We analyse the $\name$-robustness of a number of CFX approaches in the literature, demonstrating the utility of the notion and the lack of robustness in these methods (§\ref{ssec:evaluate_rob}).
    %
    %
    \item We demonstrate how the verification of $\name$-robustness can be embedded in 
    existing methods to generate CFXs which are provably robust (§\ref{ssec:generate_rob}).
\end{itemize}

We then conclude and look ahead to the various avenues of future work highlighted by our approach (§\ref{sec:conclusion}). In summary, this work presents the first approach to formally reason about and deterministically quantify CFXs' robustness to model changes in neural networks.\footnote{The code for \AR{the} implementations and experiments is publicly available at https://github.com/junqi-jiang/robust-ce-inn. 
This is the full version of the paper of the same title appearing at AAAI 2023. This version includes proofs and additional experimental details.}

\section{Related Work}
\label{sec:related}

\subsection{Approaches to CFX Generation}
\label{sec:cfxrelated}
The seminal work of \cite{Wachter_17} casts the problem of finding CFXs for neural networks as  gradient-based optimisation against the input vector using a single loss function to address the validity 
of counterfactual instances, as well as their closeness to the input instances measured by some 
distance metric (\emph{proximity})
, while that of~\cite{Tolomei_2017} defines CFXs for tree ensembles. 
Following these works,~\cite{mothilal2020explaining} include stochastic point processes and novel loss terms to generate a \emph{diverse} set of CFXs.~\cite{poyiadzi2020face} formulate the problem 
\FT{in graph-theoretic terms} and apply shortest path algorithms 
\FT{to} find CFXs that lie in the data manifold of the dataset. \cite{van2021interpretable} address the same problem using class prototypes found by variational auto-encoders 
or k-d trees.  
~\cite{MohammadiKBV21} model the generation of CFXs as a constrained optimisation problem where a neural network is encoded using Mixed-Integer Linear Programming (MILP).
Other methods that are able to generate CFXs for neural networks include that of~\AR{\cite{karimi2020model}}, which reduces CFX generation to a satisfiability problem, and that of \AR{\cite{dandl2020multi}}, which formulates 
the search for CFXs as a multi-objective optimisation problem. 
Orthogonal to these studies, 
ongoing works 
try to embed causal constraints when finding CFXs~\cite{mahajan2019preserving,karimi2021algorithmic,Kanamori_21}.
Finally, there are a number of methods for generating CFXs for linear or Bayesian models, e.g.,~\cite{ustun2019actionable,albini2020relation,Kanamori_20}, but we omit their details here since our focus is on neural networks. 

\subsection{Robustness of Models and Explanations}

Robustness has been advocated in a number of ways in AI, 
including by requiring
that outputs of neural networks should be robust to perturbations \AR{in} 
inputs~\cite{Carlini_17,Weng_18} \FL{
\AR{or in }model parameters~\cite{TsaiHYC21}}.
A number of works have drawn attention to the links between adversarial examples and CFXs, given that they solve a similar optimisation problem~\cite{Pawelczyk_22,Freiesleben_22}.
The protection which robustness to input perturbations provides against manipulation has been shown to be important 
also as concerns explanations for 
models' outputs~\cite{Slack_21} 
and a range of methods for producing explanations which are robust to input perturbations have been proposed, e.g., ~\cite{Alvarez-Melis_18,Sharma_20,Huai_22}.
Meanwhile,~\cite{Qiu_22} use input perturbations to ensure that explanations are robust to out-of-distribution data, applying this technique to a range of XAI methods for producing saliency maps.
A causal view is taken by~\cite{Hancox-Li_20} in discussing the importance of robustness to input perturbations in explanations for models' outputs. Here, it is argued that explanations should be robust to different models, not only changes within the model (as we target), if real patterns in the world are of interest. 
Ensuring that CFXs fall on the data manifold has been found to increase this robustness to multiplicity of models~\cite{Pawelczyk_22}. However, our focus is on formal approach to robustness when changing the model parameters, rather than the model itself.
Notwithstanding the findings of recent works demonstrating the significant effects of changes to model parameters on the validity of CFXs~\cite{Rawal_20X,Dutta_22}, we are aware of only two works which target the same form of robustness we consider. 
\cite{UpadhyayJL21} design a novel objective for CFXs which incorporates the model shift, i.e., the change in a model's parameters which may be, for example, weights or gradients. However, the\FT{ir} approach is heuristic and may fail to generate valid robust CFXs (we will discuss other limitations of this approach later in §\ref{ssec:evaluate_rob}). \cite{Dutta_22} define the metric of \emph{counterfactual stability}, i.e., robustness to model changes induced during retraining, before introducing an approach which refines any base method for finding CFXs in tree-based classifiers, rather than the neural networks we target.  In addition, both works evaluate CFXs' robustness by demonstrating CFXs' validity on a small number of
retrained models and cannot exhaustively prove the validity for other model changes.









\section{Preliminaries}
\label{sec:prel}

\paragraph{\textbf{Notation.}} Given an integer $k$, let $[k]$ denote the set $\{1,\ldots,k\}$. Given a set $S$, let $\card{S}$ denote its cardinality. Given a vector $x \in \mathbb{R}^n$ we use $x[i]$ to denote its $i$-th component; similarly, for a matrix $w \in \mathbb{R}^n \times \mathbb{R}^m$, we use $w[i,j]$ to denote element $i,j$. Finally, we use $\powerset{\mathbb{R}}$ to denote the set of all closed intervals over $\mathbb{R}$.



\paragraph{\textbf{Feed-forward neural networks.}} A feed-forward neural
network (FFNN) is a directed acyclic graph whose nodes are structured in layers. Formally, we describe FFNNs and the computations they perform as follows.

\begin{definition} A \emph{fully-connected feed-forward neural network (FFNN)} is a tuple $\model = (k,N,E,B,\Omega)$ where:
\begin{itemize}
    \item $k\FT{\geq 0}$ is the depth of $\model$;
    \item $(N,E)$ is a directed graph;
    \item $N = \bigsqcup_{i=0}^{k+1} N_i$ is the disjoint union of sets of nodes $N_i$;
    we call $N_0$ the input layer, $N_{k+1}$ the output layer and $N_i$ hidden layers for $i \in [k]$; 
    \item $E \!=\! \bigcup_{i=1}^{k+1}(N_{i-1} \times N_i)$ is the set of edges 
    \FT{between} layers;
    \item $B\!\!: \!\! (N \! \setminus \!\! N_0)\!\!\rightarrow\!\!\mathbb{R}$ assigns 
    bias to 
    nodes in non-input layers;
    \item $\Omega\!\!: \!\!E \rightarrow \mathbb{R}$ assigns a weight to each edge. 
\end{itemize}
\label{def:ffnn}
\end{definition}

In the following\FT{, unless specified otherwise, we assume as given an FFNN 
$\model = (k,N,E,B,\Omega)$, and}  we use 
$B_i$ to denote the vector of biases assigned to layer $N_i$ and $W_i$ to denote the matrix of weights assigned to edges between 
nodes 
\FT{in} subsequent layers $N_{i-1}, N_{i}$, for $i \in [k+1]$.

\begin{definition} Given an input $x \in \mathbb{R}^{\card{N_0}}$, an FFNN $\model$ computes an \emph{output} $\model(x)$ defined as 
follows. Let:
\begin{itemize}
    \item $V_0 = x$;
    \item $V_i = \sigma(W_{i} \cdot V_{i-1} + B_i)$  for $i \in [k]$, where $\sigma$ is an activation function applied element-wise. For $V_i = [v_{i,1}, \ldots, v_{i, \card{N_i}}]$,  $v_{i,j}$ is the \emph{value} of the $j$-th node in layer $N_i$.
    \end{itemize}
    Then, $\model(x) = V_{k+1} = W_{k+1} \cdot V_k + B_{k+1}$.
\label{def:ffnn_computation}
\end{definition}


The \emph{Rectified Linear Unit (ReLU)} activation, defined as $\sigma(x) \delequal \max(0, x)$, is perhaps the most common choice for hidden layers. We will therefore focus on FFNNs using ReLU activations in this paper.

\begin{definition} Consider an input $x  \in \mathbb{R}^{\card{N_0}}$ and an FFNN $\network$. We say that $\network$ \emph{classifies $x$ as $c$}, denoted (with an abuse of notation) $\network(x) = c$, if $c \in \argmax_{i \in [|N_{k+1}|]} \model(x)[i]$.
\label{def:ffnn_classification}
\end{definition}

For ease of exposition, §\ref{sec:robust_cfx} will focus on FFNNs used for binary classification tasks with $|N_{k+1}|=2$. The same ideas also apply to other settings, e.g., multiclass classification or binary classification using a single output node with sigmoid activation, which we use in our experiments in §\ref{sec:applications}. 

\paragraph{\textbf{Counterfactual explanations.}} Consider an FFNN $\model$ trained to solve a binary classification problem. Assume 
$\model$ produces a classification outcome $\model(x) = c$ \FT{ for input $x$}. Intuitively, a 
CFX is a new input $x'$ which is similar to $x$ and for which $\model(x') = 1 - c$. 
Formally, existing literature characterises CFXs in terms of the solution space of a Constrained Optimisation Problem (COP) as follows.

\begin{definition}\label{def:cfx} 
Consider an input $x \in \mathbb{R}^{\card{N_0}}$ and a binary classifier $\model$ s.t. $\model(x) = c$. Given a distance metric $d: \mathbb{R}^{\card{N_0}} \times \mathbb{R}^{\card{N_0}} \rightarrow \mathbb{R}$, 
a 
\emph{CFX} is any $x'$ such that:
\begin{subequations}
\begin{alignat}{2}
&\argmin_{x'}  && d(x,x')\label{eqn:obj} \\
&\text{subject to} && \quad \model(x') = 1- c, \text{ } x' \in \mathbb{R}^{\card{N_0}}\label{eqn:eq-1}
\end{alignat}
\label{eqn:all-lines}
\end{subequations}
\end{definition}

A CFX thus corresponds to the closest input $x'$ (Eq.~\ref{eqn:obj}) belonging to the original input space that makes the classification flip (Eq.~\ref{eqn:eq-1}). A common choice for the distance metric $d$ is the normalised $L_1$ distance~\cite{Wachter_17}. Under this choice, CFX generation for
FFNNs with ReLU activations can be solved exactly via MILP -- see, e.g., ~\cite{MohammadiKBV21}. Finally, we mention that the optimisation problem can also be extended to account for additional CFX properties mentioned in §\ref{sec:cfxrelated}.

We conclude with an example which summarises the main concepts presented in this section.
\begin{example}
    \label{ex:ffnn}
     Consider the FFNN $\model$ below where weights are as indicated in the diagram, biases are zero and R denotes ReLU activations.
%
%
        %
	The network receives a two-dimensional input $x = [x_0,x_1]$ and produces a two-dimensional output $y=[y_0,y_1]$. 
	
	\begin{figure}[ht]
		\centering
		\scalebox{1}{\begin{tikzpicture}

  \node[] (phantom_1) at (-1,0) {$x_0$};
  \node[] (phantom_2) at (-1,-1.5) {$x_1$};
  
  \node[circle,draw=black, minimum width=0.5cm] (input_1) at (0,0) {};
  \node[circle,draw=black, minimum width=0.5cm] (input_2) at (0,-1.5) {};
  
  \node[circle,draw=black, minimum width=0.5cm] (hidden_1) at (2,0) 
  {\scriptsize R};
  \node[circle,draw=black, minimum width=0.5cm] (hidden_2) at (2,-1.5) 
  {\scriptsize R};
  
    \node[circle,draw=black, minimum width=0.5cm] (output_1) at (4,0) 
  {};
  \node[circle,draw=black, minimum width=0.5cm] (output_2) at (4,-1.5) 
  {};
  
  \node[] (phantom_3) at (5,0) {$y_0$};
  \node[] (phantom_4) at (5,-1.5) {$y_1$};
  
  \draw[->] (phantom_1) -- (input_1);
  \draw[->] (phantom_2) -- (input_2);
  
  \draw[->] (input_1) edge node[above]{\tiny{$1$}} (hidden_1);
  \draw[->] (input_1) edge node[below, xshift=-0.7cm, yshift=-0.1cm]{\tiny$-1$} 
  (hidden_2);
  
  \draw[->] (input_2) edge node[above, xshift=-0.7cm, yshift=0.1cm]{\tiny$-1$} 
  (hidden_1);
  \draw[->] (input_2) edge node[below]{\tiny$1$} 
  (hidden_2);
  
  \draw[->] (hidden_1) edge node[above]{\tiny{$1$}} (output_1);
  \draw[->] (hidden_1) edge node[below, xshift=-0.7cm, yshift=-0.1cm]{\tiny{$0$}} (output_2);

  \draw[->] (hidden_2) edge node[above, xshift=-0.7cm, yshift=0.1cm]{\tiny$0$}  (output_1);
  \draw[->] (hidden_2) edge node[below]{\tiny$1$}  (output_2);

  \draw[->] (output_1) -- (phantom_3);

  \draw[->] (output_2) -- (phantom_4);
  
\end{tikzpicture}}
		\label{fig:original_net}
	\end{figure}
	
	The symbolic expressions for the output components 
  are $y_0 = \max(0,x_0-x_1)$ and $y_1 = \max(0, x_1 - x_0)$.  
	Given a concrete input $x=[1,2]$, we have $\model(x) = 1$. 
 \FT{A possible CFX may be $x'= [2.1,2]$, with 
 $\model(x') = 0$.}
\end{example}

\section{$\name$-Robustness via Interval Abstraction}
\label{sec:robust_cfx}
The COP formulation of CFXs presented in Definition~\ref{def:cfx} focuses on finding CFXs that are as close as possible to the original input. The rationale behind this choice is that changes in input features suggested by minimally distant CFXs likely require less effort, thus making them more easily attainable by users in real-world settings. However, it has been shown~\cite{Rawal_20X,Dutta_22} that slight changes applied to the classifier, e.g., following retraining, may impact the validity of CFXs, particularly those which are minimally distant from the original input. 
This fragility of CFXs can have troubling implications, both for the users of explanations, and for those who generate them, as discussed in §\ref{sec:intro}.

This state of affairs motivates the primary objective of this work: \textit{can we generate useful CFXs for FFNNs that are provably robust to model changes?}

In the following we formalise the notion of robustness we target and introduce an abstraction-based framework to reason about 
this notion in CFXs for FFNNs. To this end, we 
begin by defining a notion of distance between 
FFNNs.

\begin{definition}
Consider two FFNNs $\model = (k,N,E,B,\Omega)$ and $\model' = (k',N',E',B',\Omega')$. We say that $\model$ and $\model'$ \emph{have identical topology} if $k=k'$ and $(N,E) = (N',E')$.
\label{def:id_topology}
\end{definition}

\begin{definition}
Let $\model\!=\!(k,N,E,B,\Omega)$ and $\model'\!=\!(k,N,E,B',\Omega')$ be two FFNNs with identical topology. 
For $0 \!\leq \!p \! \leq\! \infty$, the \emph{$p$-distance between $\model$ and $\model'$} 
is:
\begin{equation*}
    \distance{\model}{\model'}{p} = \left( \sum_{i=1}^{k+1} \sum_{j=1}^{\card{N_i}} \sum_{
    l=1}^{\card{N_{i-1}}} \lvert W_i[j,l] - W'_i[j,l] \rvert^p \right)^{\frac{1}{p}}
\end{equation*}
\label{def:distance_between_models}
\end{definition}

Intuitively $p$-distance compares the weight matrices of $\model$ and $\model'$ and computes their distance as the $p$-norm of their difference. Biases have been omitted from Definition~\ref{def:distance_between_models} for readability; the definition can be readily extended to 
include biases too, as is the case in our implementation. Using this notion we can characterise a model shift as follows.

\begin{definition}
Given $0\!\leq \!p \!\leq \! \infty$, a \emph{model shift} is a function $\mshift$ mapping an FFNN $\model$ into another 
$\model'\!=\!\mshift(\model)$ such that:   
\begin{itemize}
    \item $\model$ and $\model'$ have identical topology;
    \item $\distance{\model}{\model'}{p} > 0$
    .
\end{itemize}
\label{def:model_shift}
\end{definition}

Model shifts are typically observed in real-world applications when a model is regularly retrained to incorporate new data. In such cases, models are likely to see only small changes at each update. In the same spirit as~\cite{UpadhyayJL21}, we capture this 
as follows.

\begin{definition}

Given an FFNN $\model$, 
\FT{$\delta \!\in\! \mathbb{R}_{>0}$ and $0 \!\leq \!p \!\leq \!\infty$},
\emph{the set of plausible model shifts} is 
$\Delta = \{ \mshift \mid \distance{\model}{\mshift(\model)}{p} \leq \delta \}$.
\label{def:set_of_plausible_shifts}
\end{definition}

Plausibility implicitly bounds the magnitude of weight and bias changes that can be effected by a model shift $\mshift$, as stated in the following.\footnote{Proofs are provided in Appendix \ref{app:proofs}}.

\begin{lemma}
Consider an FFNN $\model$ and a set of plausible model shifts $\Delta$. Let $\model' = \mshift(\model)$ for $\mshift \in \Delta$. The magnitude of weight and bias changes in $\model'$ is bounded.
\label{lemma:bounds}
\end{lemma}

In essence, it is possible to show that each weight (and bias) can change up to a maximum of $\pm \delta$ following the application of a model shift $\mshift \in \Delta$.

\begin{remark}
In the following we use $\underline{W_i'[j,l]} \delequal W_i[j,l] - \delta$ and  $\overline{W_i'[j,l]} \delequal W_i[j,l] + \delta$ to denote, respectively, the minimum and maximum value each weight $W_i'[j,l]$ can take in $\model'\FT{=\mshift(\model)}$ for any $\mshift \in \Delta$ and $i \in [k+1]$, $j\in [\card{N_i}]$ and $l \in [\card{N_{i-1}]}$. 
(Analogous notation is used for biases.)
These bounds 
are sound, but also conservative, i.e., they may result in models that exceed the upper bound on the p-distance for some choice of $p$. As an example, consider what happens when $W'_i[j,l] = \overline{W_i'[j,l]}$ for each $i \in [k+1]$, $j \in \card{N_i}$, $l \in \card{N_{i-1}}$. These valuations satisfy Definition~\ref{def:set_of_plausible_shifts} when $p=\infty$, but fail to do so for, e.g., $p=2$.
\label{remark:conservative_bounds}
\end{remark}

Despite weight changes being bounded, several different model shifts may satisfy the plausibility constraint. To guarantee robustness to model changes, one needs a way to represent and reason about the potentially infinite family of networks originated by applying each $\mshift \in \Delta$ to $\model$ compactly. We 
introduce an abstraction framework that can be used to this end. We begin by recalling the notion of \emph{interval neural networks}, as introduced in~\cite{PrabhakarA19}.

\begin{definition} An \emph{interval neural network (INN)} is a tuple $\intervalnet = (k,N,E,B_{\intervalnet},\Omega_{\intervalnet})$ where:
\begin{itemize}
    \item $k, N, E$ are as per Definition~\ref{def:ffnn};
    \item $B_{\intervalnet}: (N \setminus N_0)\rightarrow \powerset{\mathbb{R}}$ assigns 
    interval-valued bias\FT{es} to nodes in non-input layers;
    \item $\Omega_{\intervalnet}: \! E\! \rightarrow \!\powerset{\mathbb{R}}$ assigns 
    interval-valued weight\FT{s} to 
    \FT{edges}. 
\end{itemize}
\label{def:inn}
\end{definition}

\begin{example}
\label{ex:intervalNN}
	The diagram below shows an example of an INN. As we can observe, the INN differs from a standard FFNN in that weights and biases are intervals.
	\begin{figure}[ht]
		\centering
		\scalebox{1}{\begin{tikzpicture}

  \node[] (phantom_1) at (-1,0) {$x_0$};
  \node[] (phantom_2) at (-1,-1.5) {$x_1$};
  
  \node[circle,draw=black, minimum width=0.5cm] (input_1) at (0,0) {};
  \node[circle,draw=black, minimum width=0.5cm] (input_2) at (0,-1.5) {};
  
  \node[circle,draw=black, minimum width=0.5cm] (hidden_1) at (2,0) 
  {\scriptsize R};
  \node[circle,draw=black, minimum width=0.5cm] (hidden_2) at (2,-1.5) 
  {\scriptsize R};
  
      \node[circle,draw=black, minimum width=0.5cm] (output_1) at (4,0) 
  {};
  \node[circle,draw=black, minimum width=0.5cm] (output_2) at (4,-1.5) 
  {};
  
  \node[] (phantom_3) at (5,0) {$y_0$};
  \node[] (phantom_4) at (5,-1.5) {$y_1$};
  
  \draw[->] (phantom_1) -- (input_1);
  \draw[->] (phantom_2) -- (input_2);
  
  \draw[->] (input_1) edge node[above]{\tiny{$[0.9.1.1]$}} (hidden_1);
  \draw[->] (input_1) edge node[below, xshift=-1.2cm, yshift=-0.1cm]{\tiny$[-1.1,-0.9]$} 
  (hidden_2);
  
  \draw[->] (input_2) edge node[above, xshift=-1.2cm, yshift=0.1cm]{\tiny$[-1.1,-0.9]$} 
  (hidden_1);
  \draw[->] (input_2) edge node[below]{\tiny$[0.9,1.1]$} 
  (hidden_2);
  
\draw[->] (hidden_1) edge node[above]{\tiny{$[0.9.1.1]$}} (output_1);
  \draw[->] (hidden_1) edge node[below, xshift=1.1cm, yshift=-0.1cm]{\tiny{$[-0.1,0.1]$}} (output_2);

  \draw[->] (hidden_2) edge node[above, xshift=1.1cm, yshift=0.1cm]{\tiny$[-0.1,0.1]$}  (output_1);
  \draw[->] (hidden_2) edge node[below]{\tiny$[0.9.1.1]$}  (output_2);
  
  \draw[->] (output_1) -- (phantom_3);

  \draw[->] (output_2) -- (phantom_4);
  
\end{tikzpicture}}
		\label{fig:abst_net}
	\end{figure}
	
\end{example}


\FT{In the remainder, unless specified otherwise, when using an INN we will assume its components are as in Definition~\ref{def:inn}.} 
We will use boldface notation to denote interval-valued biases $\mathbf{B}_i$ and weights $\mathbf{W}_i$.
The computation performed by an INN differs from that of an FFNN as follows.

\begin{definition} Given an input $x \in \mathbb{R}^{\card{N_0}}$, an INN $\intervalnet$ computes an \emph{output} $\intervalnet(x)$ defined as follows. Let:
\begin{itemize}
    \item $\mathbf{V}_0 = [x,x]$;
    \item $\mathbf{V}_i = \mathbf{\sigma}(\mathbf{W}_i \cdot \mathbf{V}_{i-1} + \mathbf{B}_i)$  for $i \in [k]$. For $\mathbf{V}_i = [\mathbf{v}_{i,1},\ldots,\mathbf{v}_{i,\card{N_i}}]$,  $\mathbf{v}_{i,j} = [v_{i,j}^l, v_{i,j}^u ]$ is the \emph{interval of values} for the $j$-th node in layer $N_i$.
\end{itemize}
Then, $\intervalnet(x) = \mathbf{V}_{k+1}  = \mathbf{W}_{k+1} \cdot \mathbf{V}_k + \mathbf{B}_{k+1}$.
\label{def:inn_computation}
\end{definition}

\FT{Thus,} an INN computes an interval for each output node. These intervals contain all possible values that each output can take under the valuations induced by $B_{\intervalnet}$ and $\Omega_{\intervalnet}$. As a result, 
the classification semantics of an INN \FT{is} as follows.




\begin{definition} Consider an input $x\in \mathbb{R}^{\card{N_0}}$, a binary label $c$ and an INN $\intervalnet$. We say that
 $\intervalnet$ 
    \emph{classifies $x$ as $c$}, written $\intervalnet(x) = c$, if ${v}_{k+1,c}^l > {v}_{k+1,1-c}^u$. 
\label{def:inn_classification}
\end{definition}

%
%
%
%

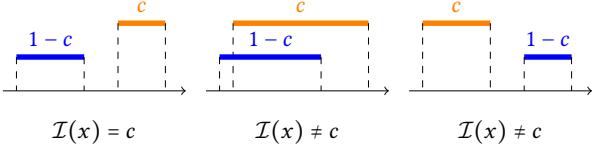
\begin{figure}
    \centering
    \scalebox{0.9}{\begin{tikzpicture}

 
  
 

  

  

\draw[line width=0.8mm, color=orange] (4,0) edge node[above]{$c$} (4.7,0);
\draw[dashed] (4,0) -- (4,-1);
\draw[dashed] (4.7,0) -- (4.7,-1);

\draw[line width=0.8mm,color=blue] (2.5,-0.5) edge node[above]{$1-c$} (3.5,-0.5);
\draw[dashed] (2.5,-0.5) -- (2.5,-1);
\draw[dashed] (3.5,-0.5) -- (3.5,-1);

\draw[->] (2.3,-1) -- (5,-1);
\node[] (phantom_r1) at (4.9,-1.3) {}
;

\node[] (phantom_1) at (3.65,-1.6) {$\mathcal{I}(x) = c$};


\draw[line width=0.8mm, color=orange] (5.7,0) edge node[above]{$c$} (7.7,0);
\draw[dashed] (5.7,0) -- (5.7,-1);
\draw[dashed] (7.7,0) -- (7.7,-1);

\draw[line width=0.8mm,color=blue] (5.5,-0.5) edge node[above, xshift=0.0cm]{$1-c$} (7.,-0.5);
\draw[dashed] (5.5,-0.5) -- (5.5,-1);
\draw[dashed] (7,-0.5) -- (7,-1);

\draw[->] (5.3,-1) -- (8,-1);
\node[] (phantom_r2) at (7.9,-1.3) {};

\node[] (phantom_1) at (6.65,-1.6) {$\mathcal{I}(x) \neq c$};


\draw[line width=0.8mm, color=blue] (10,-0.5) edge node[above]{$1- c$} (10.7,-0.5);
\draw[dashed] (10,-0.5) -- (10,-1);
\draw[dashed] (10.7,-0.5) -- (10.7,-1);

\draw[line width=0.8mm,color=orange] (8.5,0) edge node[above]{$c$} (9.5,0);
\draw[dashed] (8.5,0) -- (8.5,-1);
\draw[dashed] (9.5,0) -- (9.5,-1);

\draw[->] (8.3,-1) -- (11,-1);
\node[] (phantom_r3) at (10.9,-1.3) {};

\node[] (phantom_1) at (9.65,-1.6) {$\mathcal{I}(x) \neq c$};

\end{tikzpicture}}
    \caption{Graphical 
    comparison of output intervals for class $c$ and class $1-c$, for Definition~\ref{def:inn_classification}
    . When $\intervalnet(x) = c$, the output range for class $c$ is always greater than that of class $1-c$. Otherwise, we say $\intervalnet(x) \neq c$.
    }
    \label{fig:inn_classfication}
\end{figure}

Figure~\ref{fig:inn_classfication} provides a graphical representation of this 
classification semantics
.
Using the INN as a computational backbone, we can now define the interval abstraction of an FFNN which is central to this work.

\begin{definition} Consider an FFNN $\model$ and a set of plausible model shifts $\Delta$. We define the \emph{interval abstraction of $\model$ under $\Delta$} as the interval neural network $\abst{\model}{\Delta}$ such that
$\model$ and $\abst{\model}{\Delta}$ have identical topology and \FT{the interval-valued biases/weights of $\abst{\model}{\Delta}$ are}:
\begin{itemize}
    \item 
    $\mathbf{B}_i[j]  = [\underline{B_i'[j]}, \overline{B_i'[j]}]$  for $i \in [k+1]$ and $j \in [\card{N_i}]$;
    \item  
    $\mathbf{W}_i[j,l]  = [\underline{W_i'[j,l]}, \overline{W_i'[j,l]}]$  for $i \in [k+1]$, $j \in [\card{N_i}]$ and $l \in [\card{N_{i-1}}]$.
\end{itemize}
\label{def:interval_abstraction}
\end{definition}

\begin{lemma}
$\abst{\model}{\Delta}$ over-approximates the set of models $\model'$ that can be obtained from $\model$ via $\Delta$. 
\label{lemma:over-approx}
\end{lemma}

Lemma~\ref{lemma:over-approx} states that $\abst{\model}{\Delta}$ contains all models that can be obtained from $\Delta$ and possibly more, but not 
fewer (see Appendix~\ref{app:proofs} for proofs)
. For some values of $p$, the interval abstraction 
\FT{may cease} to be an over-approximation and encode exactly 
the models that can be obtained from $\model$ via $\Delta$, e.g., when $p=\infty$.

A model shift $\mshift$, although plausible, may result in changes to the classification of the original input $x$. When this happens, robustness \FT{of explanations} becomes vacuous. 
\FT{For this reason, we will focus on \emph{sound shifts}, formulated as follows.}

\begin{definition} Consider an input $x\in \mathbb{R}^{\card{N_0}}$ and an FFNN 
$\model$ s.t. $\model(x) = c$. Let 
$\abst{\model}{\Delta}$ be the interval abstraction of $\model$ under a set of plausible model shifts
$\Delta$. We say that \emph{$\Delta$ is sound} if
$\abst{\model}{\Delta}(x) = c$.
\label{def:sound_delta}
\end{definition}

We are now ready to formally define the CFX robustness property that we target in this work.

\begin{definition} Consider an input $x\in \mathbb{R}^{\card{N_0}}$ and an FFNN 
$\model$ s.t. $\model(x) = c$. Let 
$\abst{\model}{\Delta}$ be the interval abstraction of $\model$ under a sound set of plausible model shifts
$\Delta$. We say that \emph{a CFX $x'$ is $\name$-robust} iff $\abst{\model}{\Delta}(x') = 1- c$.
\label{def:delta_robustness}
\end{definition}


We illustrate these concepts in the following example.

\begin{example}
\label{ex:abstract}  We observe that the INN in Example~\ref{ex:intervalNN} corresponds to the interval abstraction $\abst{\model}{\Delta}$ of the FFNN $\model$ 
in Example~\ref{ex:ffnn}, obtained for $\Delta = \{\mshift \mid \distance{\model}{\mshift(\model)}{\infty} \leq 0.1\}$.
	
	The symbolic expressions for the outputs of the INN are:
	\begin{equation*}
	\begin{split}
	    y_0 & = [0.9,1.1] \cdot \max(0,[0.9,1.1]\cdot x_0 + [-1.1,-0.9]\cdot x_1) + \\
	    & [-0.1,0.1] \cdot \max(0,[-1.1,-0.9]\cdot x_0 + [0.9,1.1]\cdot x_1)
	\end{split}
	\end{equation*}
	\begin{equation*}
	\begin{split}
	    y_1 & = [0.9,1.1] \cdot \max(0,[0.9,1.1]\cdot x_1 + [-1.1,-0.9]\cdot x_0) + \\
	    & [-0.1,0.1] \cdot \max(0,[-1.1,-0.9]\cdot x_1 + [0.9,1.1]\cdot x_0)
	\end{split}
	\end{equation*}
	
	Given a concrete input $x=[1,2]$, we observe that $\abst{\model}{\Delta}(x) = 1$ and thus establish that $\Delta$ is sound. We now check if the old CFX $x'= [2.1,2]$ is still valid under the model shifts captured by $\Delta$. The 
 INN outputs $y_0 = [-0.031,0.592]$ and $y_1 = [-0.051,0.392]$, indicating that $\abst{\model}{\Delta}(x') \simeq 0$. We thus conclude that $x'$ is not $\name$-robust.
	
	Assume now a different CFX $x'' = [2.6,2]$ is computed. The outputs of $\abst{\model}{\Delta}$ for $x''$ are $y_0 = [0.126,1.166]$ and $y_1 = [-0.106, 0.106]$. Since $y_0^{l} > y_1^{u}$, we have $\abst{\model}{\Delta}(x'') = 0$, proving that the new CFX is $\name$-robust.
\end{example}



As shown in Example~\ref{ex:abstract}, the interval abstraction $\abst{\model}{\Delta}$ can be used to prove whether a given CFX $x'$ is $\name$-robust. Indeed, when $\abst{\model}{\Delta}(x)=c$, we can conclude that the classification of $x'$ will remain unchanged for all $\mshift$ in $\Delta$. Checking Definition~\ref{def:inn_classification} requires the computation of the output reachable intervals for each output of the INN; for ReLU-based FFNNs, we use the MILP formulation of~\cite{PrabhakarA19} (see Appendix~\ref{app:INN}). 


\section{$\name$-Robustness in Action}
\label{sec:applications}
In §\ref{sec:robust_cfx} we
laid the theoretical foundations of an abstraction framework based on INNs that allows to  
reason 
about the robustness of CFXs compactly. In this section we demonstrate the utility thereof by considering two distinct applications:
\begin{itemize}
    \item in §\ref{ssec:evaluate_rob}, we show how the interval abstraction can be used to \textbf{analyse} the $\name$-robustness of different CFX algorithms across model shifts of increasing magnitudes;
    \item in §\ref{ssec:generate_rob}, we propose an 
    algorithm that uses interval abstractions to \textbf{generate} provably robust CFXs.
\end{itemize}

Our experiments, conducted on both homogeneous (continuous features) and heterogeneous (mixed continuous and discrete features) data types (see §\ref{ssec:data}), show that our approach provides a measure for assessing the robustness of CFXs generated by other \FT{SOTA} methods, but 
it can also be used to devise algorithms for generating CFXs with \emph{provable} robustness guarantees, in contrast with existing methods.

\subsection{Experimental Setup}
\label{ssec:data}
We consider four datasets with a mixture of heterogeneous and continuous data. We refer to them as \textit{credit} (heterogeneous)~\cite{Dua2019}, \textit{small business administration 
(SBA)}
(using only their continuous features)~\cite{studentDS}, \textit{diabetes} (continuous)~\cite{smith1988using} and \textit{no2} (continuous)~\cite{OpenML2013}.

The first two datasets contain known distribution shifts \cite{UpadhyayJL21}. We use $\dataset_1$ ($\dataset_2$) to denote the dataset before (respectively after) the shift. For the other datasets, we randomly shuffle the instances and separate them into two halves, again denoted as $\dataset_1$ and $\dataset_2$. For each dataset, we use $\dataset_1$ to train a base model, and use instances in $\dataset_2$ to generate model shifts via incremental retraining. We use $p=\infty$ in all experiments that follow.

CFXs are generated using the following SOTA algorithms. We consider \textit{Wachter et al.}~\cite{Wachter_17} (continuous data only), \textit{Proto}~\cite{van2021interpretable} and a method inspired by~\cite{MohammadiKBV21}. The first two implement CFX search via gradient descent, while the third uses 
\FT{MILP, and is thus referred to here as \textit{MILP}}. We also include \textit{ROAR}~\cite{UpadhyayJL21}, a SOTA framework \FT{specifically designed} to generate robust CFXs. More details about our experimental setup can be found in Appendix~\ref{app:setup}.

\begin{figure*}[ht!]
\centering
\begin{subfigure}{1.4\columnwidth}
    \includegraphics[width=\linewidth]{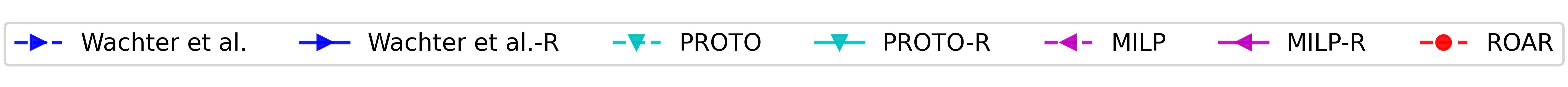}
  \end{subfigure}

  \begin{subfigure}{0.5\columnwidth}
    \includegraphics[width=\linewidth]{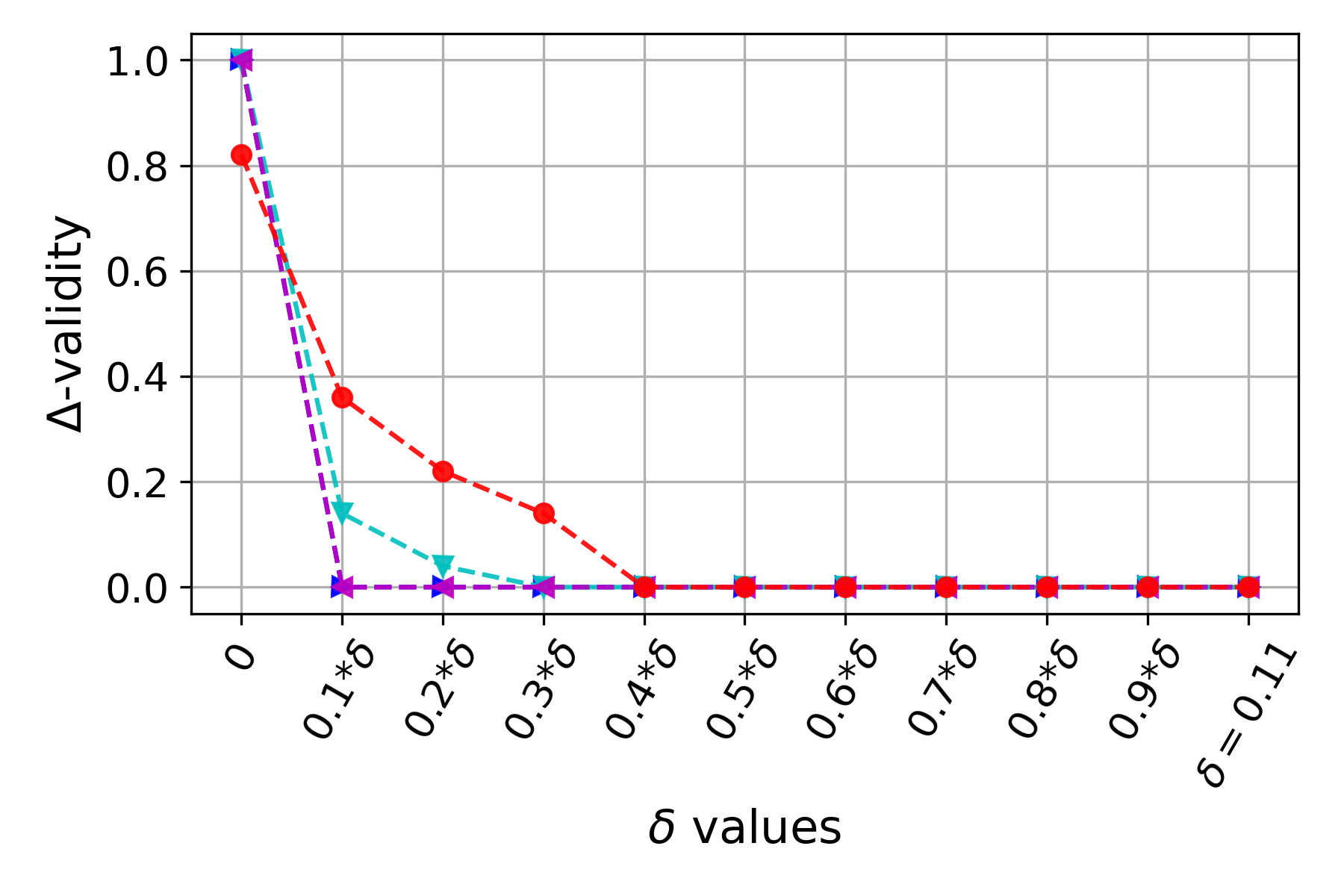}
    \caption{SOTA algorithms, diabetes}
    \label{fig:diab}
  \end{subfigure}
  \begin{subfigure}{0.5\columnwidth}
    \includegraphics[width=\linewidth]{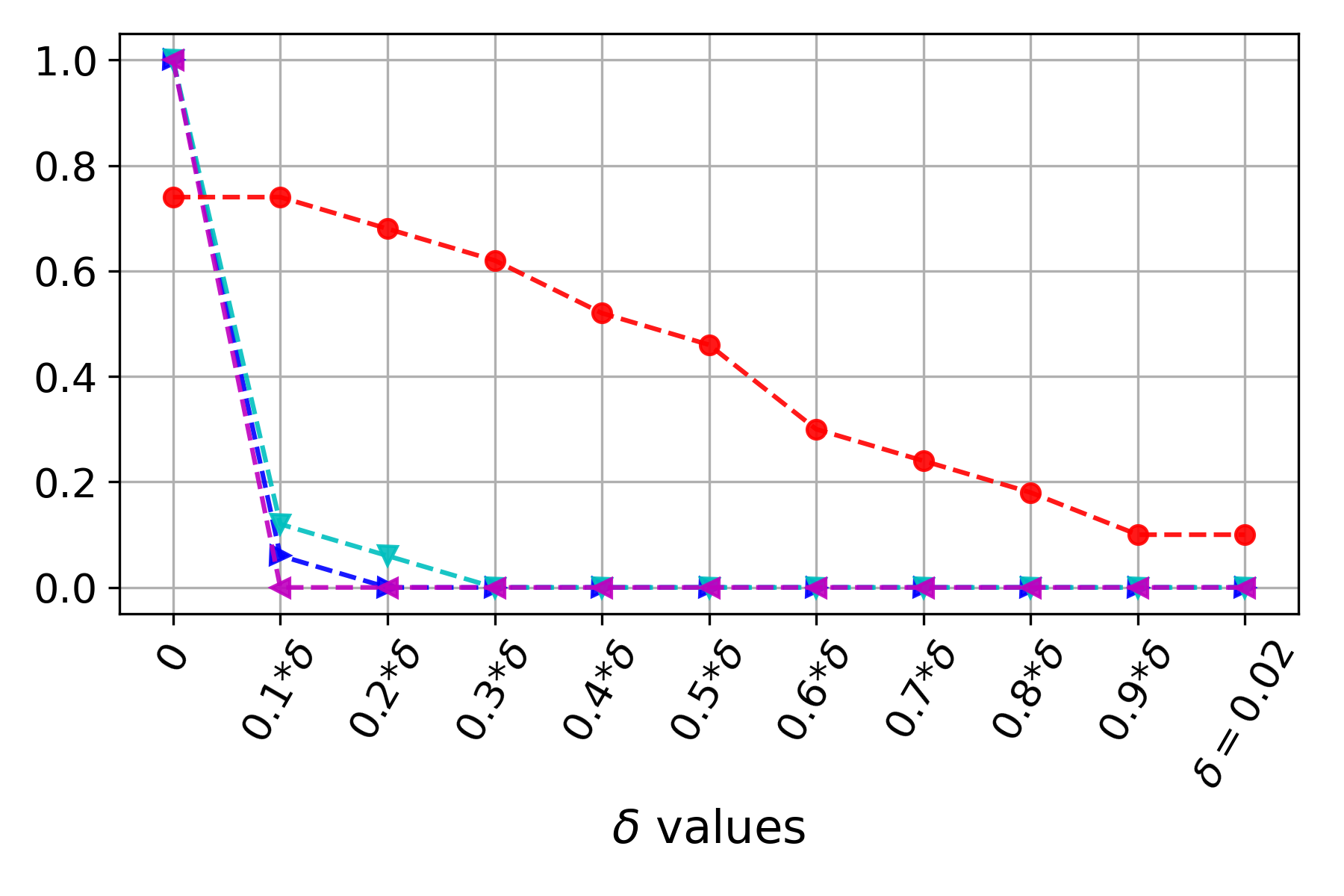}
    \caption{SOTA algorithms, no2}
    \label{fig:no2}
  \end{subfigure}
  \begin{subfigure}{0.5\columnwidth}
    \includegraphics[width=\linewidth]{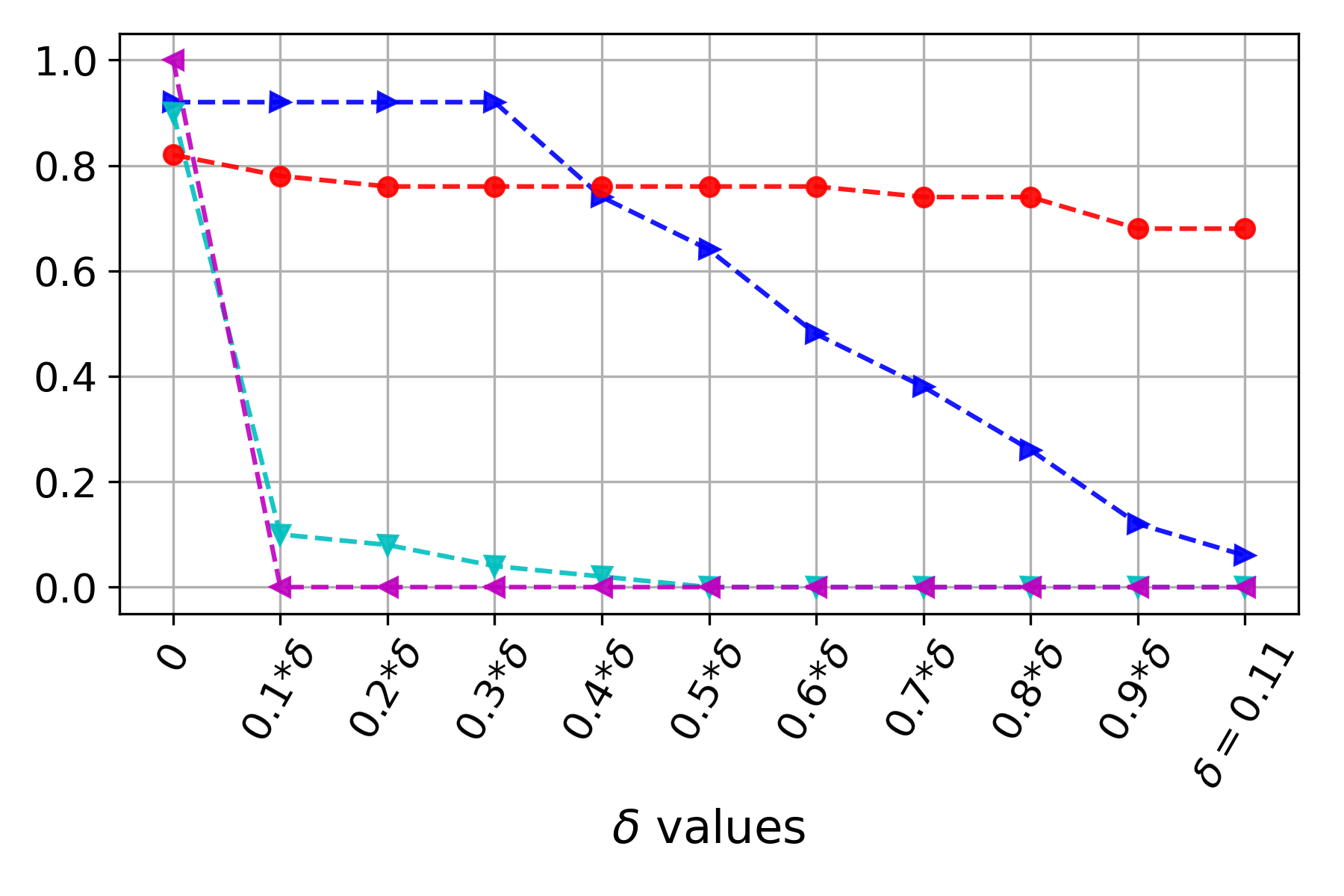}
    \caption{SOTA algorithms, SBA}
    \label{fig:sba}
  \end{subfigure}
  \begin{subfigure}{0.5\columnwidth}
    \includegraphics[width=\linewidth]{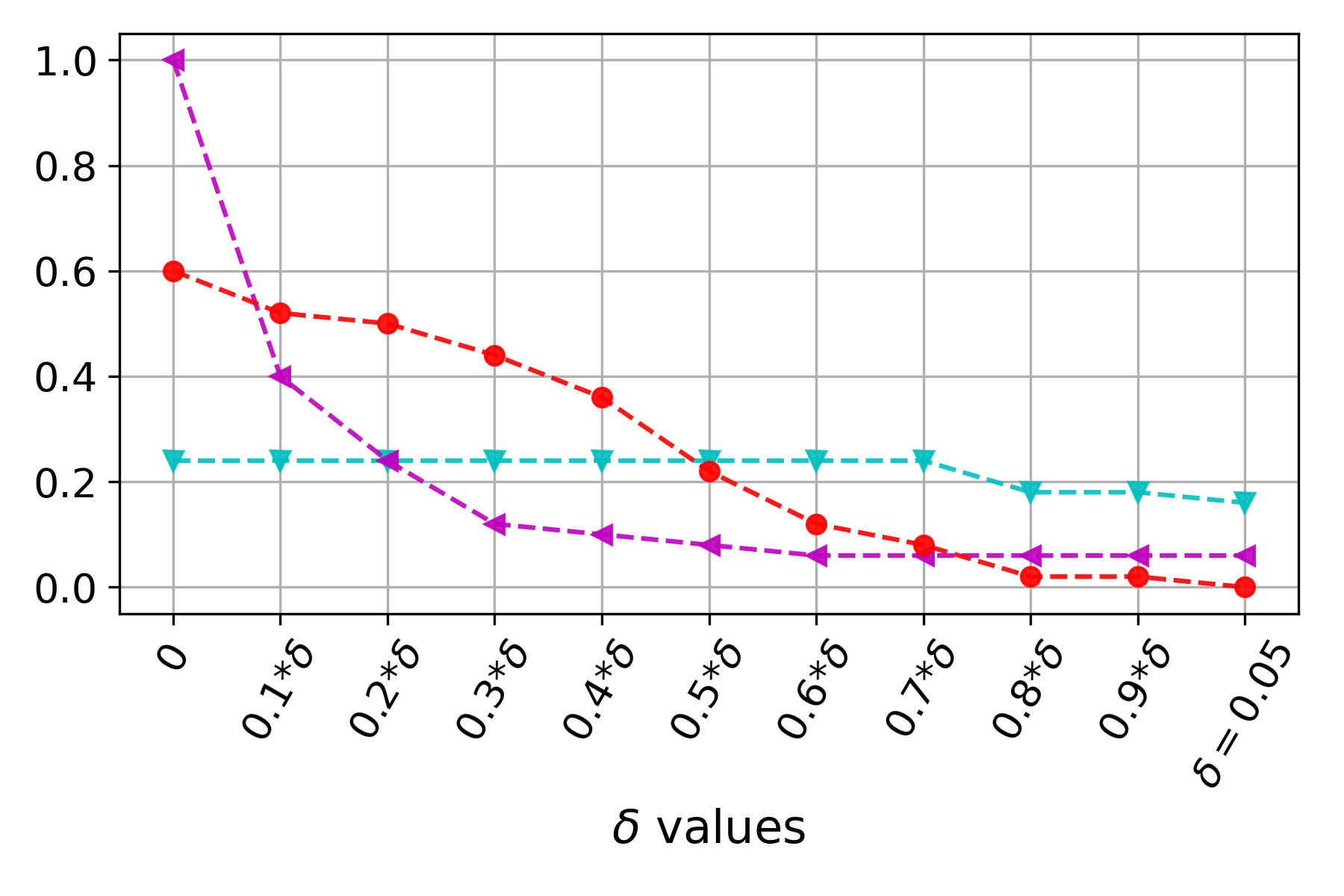}
    \caption{SOTA algorithms, credit}
    \label{fig:cred}
  \end{subfigure}
  
  \vspace{0.5cm}
  
    \begin{subfigure}{0.5\columnwidth}
    \includegraphics[width=\linewidth]{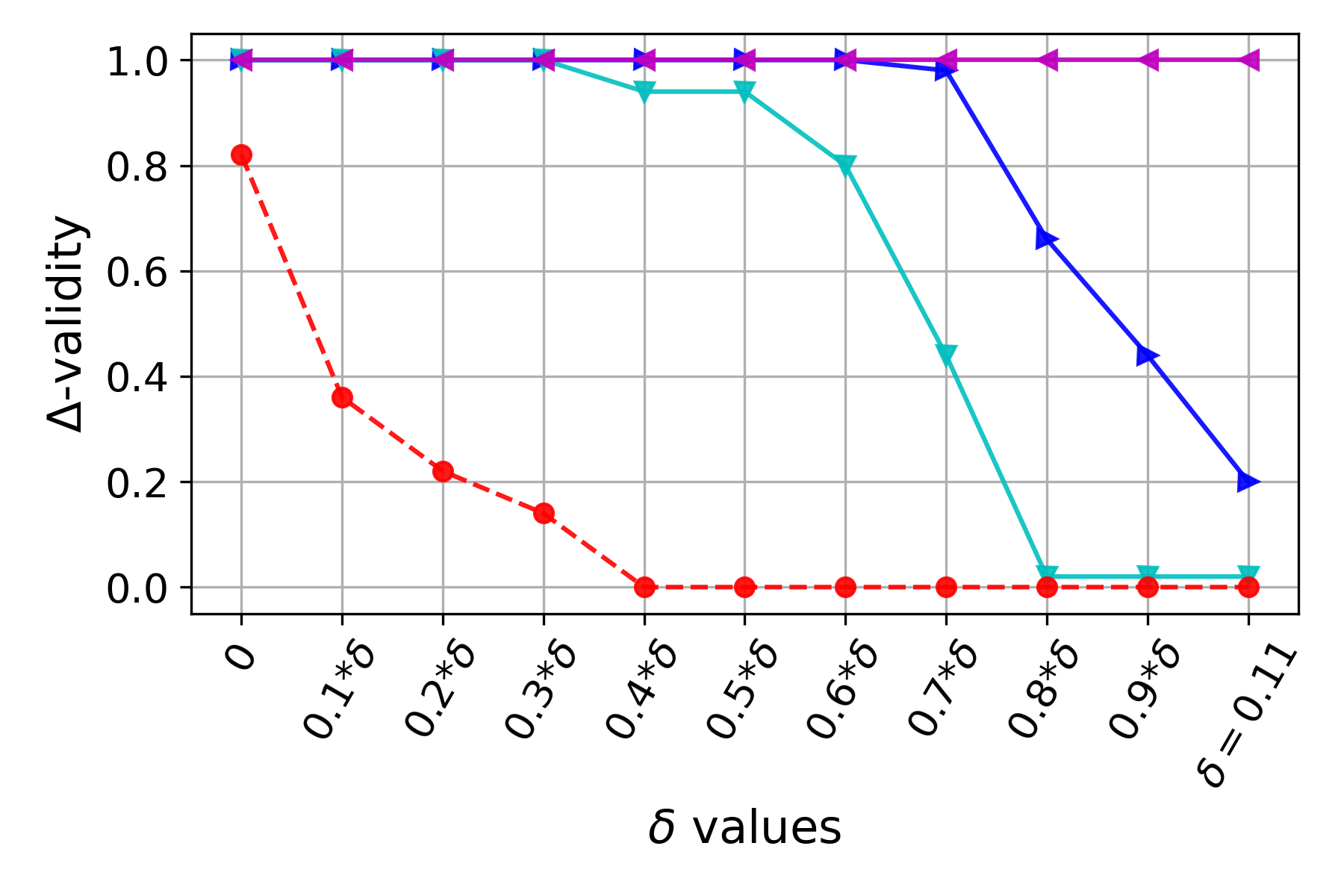}
    \caption{robust algorithms, diabetes}
    \label{fig:diabr}
  \end{subfigure}
  \begin{subfigure}{0.5\columnwidth}
    \includegraphics[width=\linewidth]{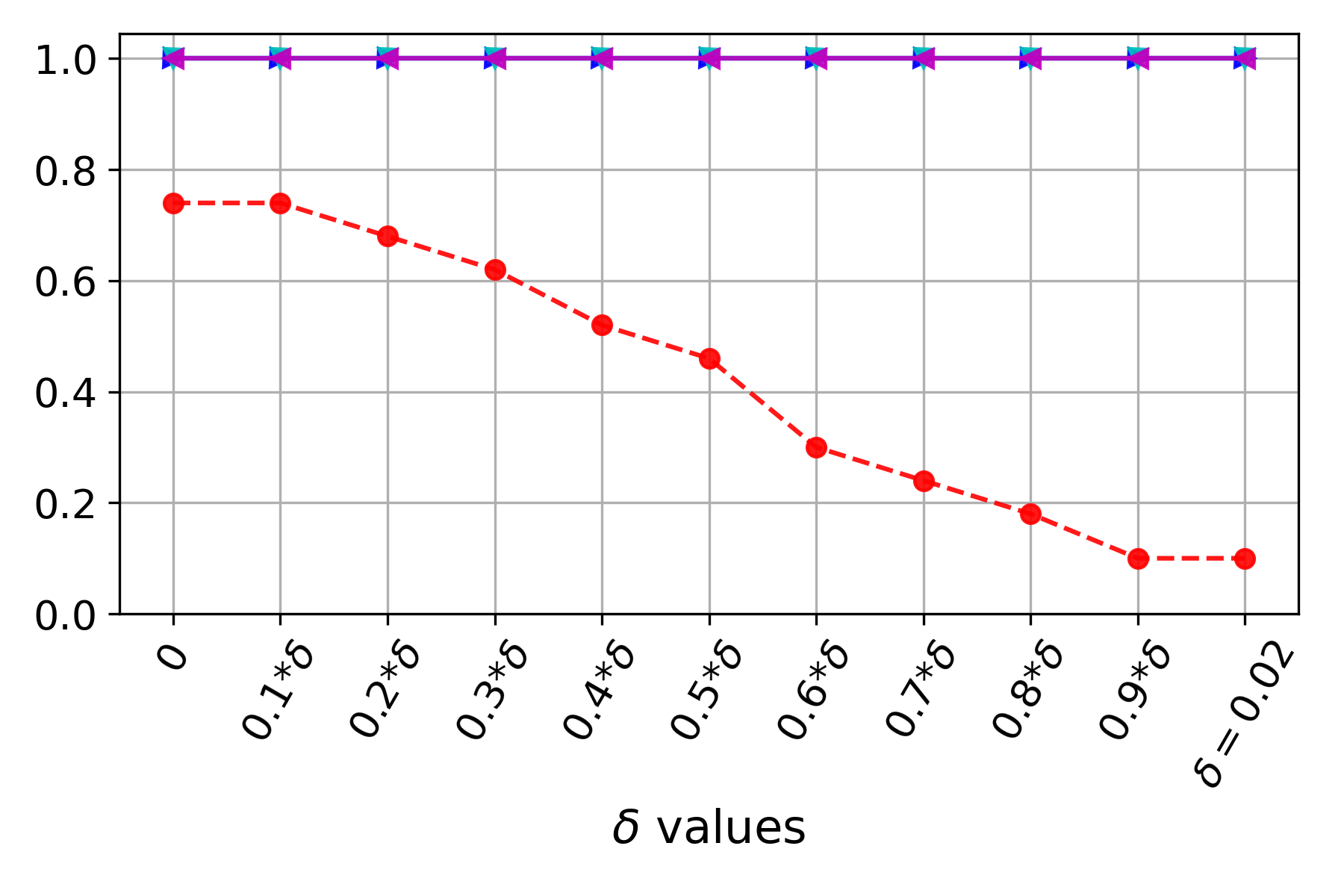}
    \caption{robust algorithms, no2}
    \label{fig:no2r}
  \end{subfigure}
  \begin{subfigure}{0.5\columnwidth}
    \includegraphics[width=\linewidth]{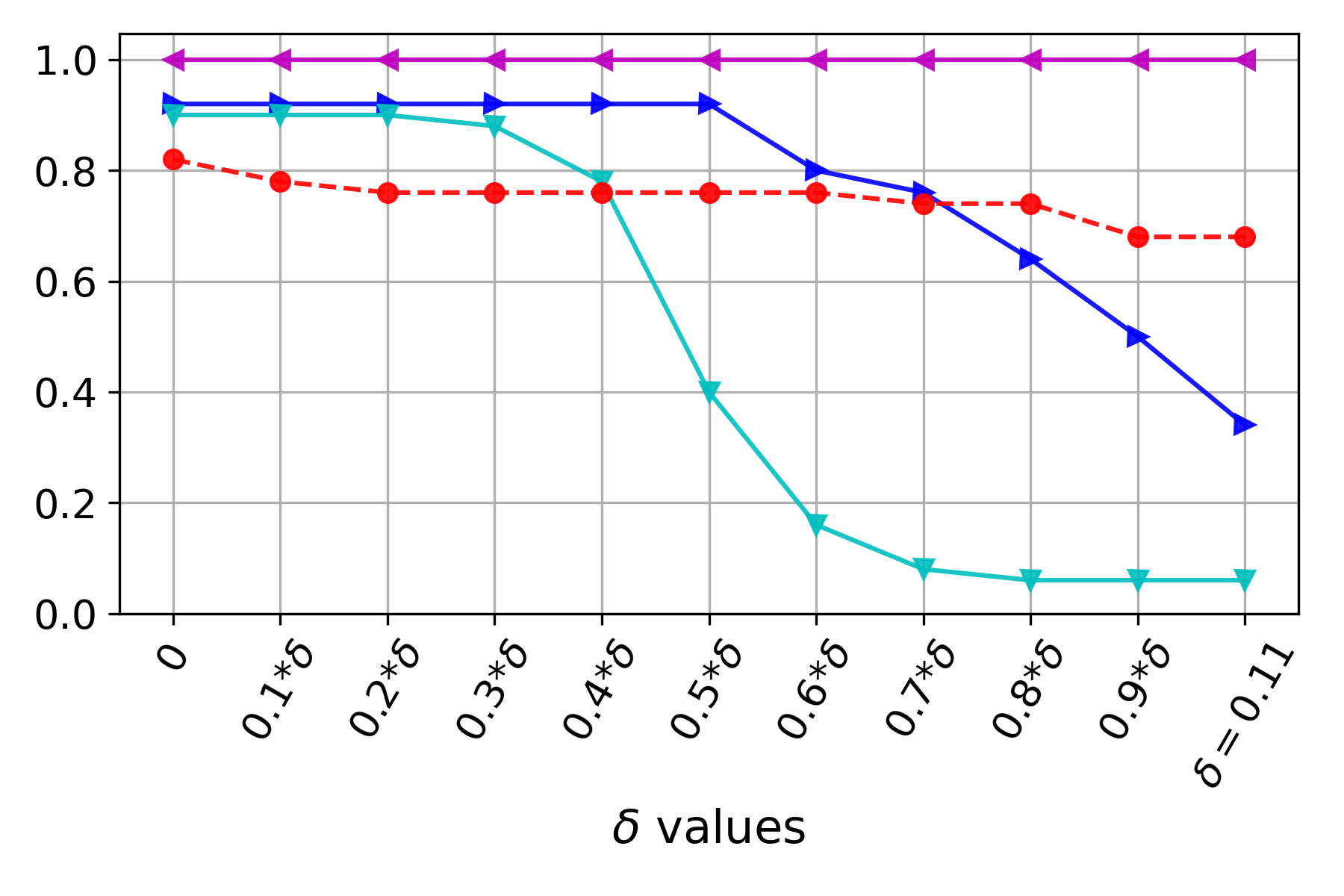}
    \caption{robust algorithms, SBA}
    \label{fig:sbar}
  \end{subfigure}
  \begin{subfigure}{0.5\columnwidth}
    \includegraphics[width=\linewidth]{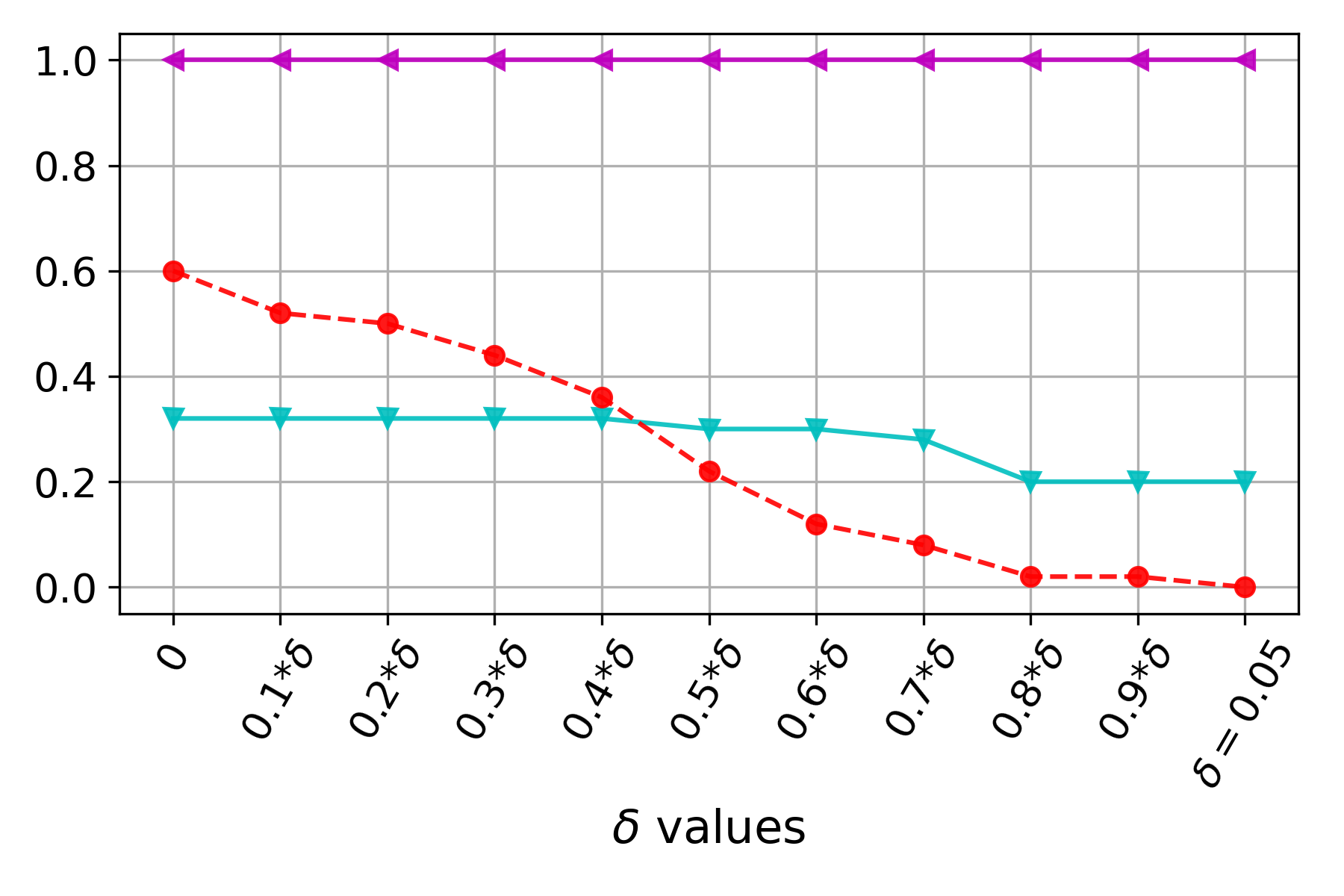}
    \caption{robust algorithms, credit}
    \label{fig:credr}
  \end{subfigure}

  \caption{Evaluation of $\name$-validity. (a-d, see §\ref{ssec:evaluate_rob}): SOTA algorithms fail to generate completely robust CFXs as 
  $\delta$ increases. (e-h, see §\ref{ssec:generate_rob}): Embedding $\name$-robustness in the search process of the same 
  algorithms results in more provably robust CFXs
  .}
    \label{fig:analysis_sota}
\end{figure*}

\subsection{Analysing $\Delta$-Robustness of CFXs}
\label{ssec:evaluate_rob}

This experiment is designed to show that interval abstractions can provide an effective tool to analyse CFXs generated by SOTA algorithms. For each dataset, we identify the largest $\delta_{max}$ that results in a set 
$\Delta$ that is sound for at least 50 test instances in $\dataset_1$.
This is 
achieved by retraining the base model using increasingly large portions of $\dataset_2$.\footnote{In real-world applications, 
values of $\delta$ could be empirically estimated by model developers by observing  retraining histories and calculating the $p$-distances between subsequent 
retraining steps.} 
We then use the CFX generation algorithms to produce $50$ CFXs. Again, see Appendix~\ref{app:setup} for details of both steps. 
 We evaluate their robustness for model shifts of magnitude up to $\delta_{max}$ using \textbf{$\Delta$-validity}, the percentage of test instances whose CFXs are $\Delta$-robust. 

Figures~\ref{fig:analysis_sota}(a-d) report the results of our analysis \FT{for the four datasets}. As we can observe, all methods generate CFXs that tend to be valid \FT{counterfactuals} for the original model ($\delta=0$), with ROAR having lower results in most cases. This is because ROAR approximates the local behaviours of FFNNs using LIME~\cite{ribeiro2016should}, which may cause a slight decrease in the \FT{counterfactual} validity~\cite{UpadhyayJL21}.
However, the picture changes as soon as small model shifts are applied. The $\Delta$-validity \FT{values} of Wachter et al., Proto and MILP quickly drop 
to zero even for model shifts of magnitude equal to $10\%$ of $\delta_{max}$, revealing that these algorithms are prone to generating non-robust CFXs 
when even very small shifts are seen in the model parameters.
ROAR exhibits a higher degree of $\name$-robustness, as expected. However, its heuristic nature does not allow to reason about all possible shifts in $\Delta$, which clearly affects the $\name$-robustness of CFXs as $\delta$ grows larger.

All methods considered here (Wachter et al, Proto, MILP and ROAR) return a single CFX for each input. However, $\name$-robustness can also be used with methods generating multiple CFXs, e.g., as with the DiCE method of \cite{mothilal2020explaining}. In these latter cases, $\name$-robustness can  be deployed  as a filter, with customisable coarseness achieved by varying $\name$, to obtain sets of \emph{diverse} and $\name$-robust CFXs. When doing so, experiments show a similar decrease of $\name$-validity as in Figure~\ref{fig:analysis_sota}(a-d).
\AR{We demonstrate this application of $\name$-robustness in Appendix~\ref{app:DiCE} and leave further exploration to future work.}

\begin{table*}[ht]
    \centering
    \resizebox{2\columnwidth}{!}{
    \begin{tabular}{ccccccccccccccccc}

        \cline{2-17}
        &
         \multicolumn{4}{c}{diabetes, target $\delta=0.11$} &
         \multicolumn{4}{c}{no2, target $\delta=0.02$} &
         \multicolumn{4}{c}{SBA, target $\delta=0.11$} &
         \multicolumn{4}{c}{\!\!\!\! credit, target $\delta=0.05$\!\!\!\!} \\
        
        \cline{2-17}
        & 
        \textbf{vm1}& 
        \!\!\!\textbf{vm2}\!\!\! & 
        \!\!\!$\ell_1$\!\!\! &
        \textbf{lof} &
        \textbf{vm1}& 
        \!\!\!\textbf{vm2}\!\!\! & 
        \!\!\!$\ell_1$\!\!\! &
        \textbf{lof} &
        \textbf{vm1}& 
        \!\!\!\textbf{vm2}\!\!\! & 
        \!\!\!$\ell_1$\!\!\! &
        \textbf{lof} &
        \textbf{vm1}& 
        \!\!\!\textbf{vm2}\!\!\! & 
        \!\!\!$\ell_1$\!\! &
        \textbf{lof} \\
        
        \hline
        \!\!\!\!Wachter et al.\!\!\!\! & 
        100\% &
        \!\!\!0\%\!\!\! &
        \!\!\!0.051\!\!\! &
        0.96 &
        100\% & 
        \!\!\!32\%\!\!\! &
        \!\!\!0.035\!\!\! &
        1.00 &
        92\%&
        \!\!\!92\%\!\!\!&
        \!\!\!0.018\!\!\!&
        -0.57 &
        - &
        - &
        - &
        - \\
        \!\!\!\!\textbf{Wachter et al.-R}\!\!\!\! & 
        100\% &
        \!\!\!100\%\!\!\! &
        \!\!\!0.122\!\!\! &
        1.00 &
        100\% &
        \!\!\!100\%\!\!\! &
        \!\!\!0.084\!\!\! &
        1.00 &
        92\%&
        \!\!\!92\%\!\!\!&
        \!\!\!0.023\!\!\!&
        -0.78&
        - &
        - &
        - &
        - \\
        \hline
        Proto & 
        100\% &
        \!\!\!18\%\!\!\! &
        \!\!\!0.063\!\!\! &
        1.00 &
        100\% &
        \!\!\!32\%\!\!\! &
        \!\!\!0.036\!\!\! &
        1.00 &
        90\%&
        \!\!\!6\%\!\!\!&
        \!\!\!0.008\!\!\!&
        0.60&
        24\%&
        \!\!\!22\%\!\!\!&
        \!\!\!0.313\!\!\!&
        -1.00\\ 
        \textbf{Proto-R} & 
        100\% &
        \!\!\!96\%\!\!\! &
        \!\!\!0.104\!\!\! &
        1.00 &
        100\%&
        \!\!\!100\%\!\!\! &
        \!\!\!0.069\!\!\! &
        1.00 &
        90\%&
        \!\!\!88\%\!\!\!&
        \!\!\!0.011\!\!\!&
        -0.02 &
        32\%&
        \!\!\!30\%\!\!\!&
        \!\!\!0.300\!\!\!&
        -1.00\\ 
        \hline
        MILP & 
        100\% &
        \!\!\!0\%\!\!\! &
        \!\!\!0.049\!\!\! &
        0.96 &
        100\%& 
        \!\!\!32\%\!\!\! & 
        \!\!\!0.032\!\!\! &
        1.00 &
        100\%&
        \!\!\!4\%\!\!\!&
        \!\!\!0.007\!\!\!&
        0.56 &
        100\%&
        \!\!\!74\%\!\!\!&
        \!\!\!0.024\!\!\!&
        1.00\\
        \textbf{MILP-R} & 
        100\% &
        \!\!\!100\%\!\!\! &
        \!\!\!0.212\!\!\! &
        -0.48 &
        100\%&
        \!\!\!100\%\!\!\! &
        \!\!\!0.059\!\!\! &
        1.00&
        100\%&
        \!\!\!100\%\!\!\!&
        \!\!\!0.018\!\!\!&
        -0.88 &
        100\%&
        \!\!\!100\%\!\!\!&
        \!\!\!0.031\!\!\!&
        1.00\\
        \hline
        ROAR & 
        82\% &
        \!\!\!14\%\!\!\! &
        \!\!\!0.078\!\!\! &
        0.95 &
        88\%&
        \!\!\!34\%\!\!\! &
        \!\!\!0.074\!\!\! &
        1.00&
        82\%&
        \!\!\!78\%\!\!\!&
        \!\!\!0.031\!\!\!&
        -0.80&
        62\%&
        \!\!\!60\%\!\!\!&
        \!\!\!0.047\!\!\!&
        1.00\\
        \hline

    \end{tabular}
    }
    \caption{Evaluating the robustness of CFXs for base methods and their $\name$-robust variants.}
    \label{tab:results-NN}
\end{table*}

\subsection{Generating Provably Robust CFXs}
\label{ssec:generate_rob}

Our earlier experiments reveal that SOTA algorithms, including those that are designed to be robust, often fail to generate CFXs that satisfy $\name$-robustness. Thus, the problem of generating CFXs that are provably robust against model shifts remains largely unsolved.
%
We will now show how $\name$-robustness can be used to guide CFX generation algorithms toward 
\FT{CFXs} with formal robustness guarantees.
Our proposed approach, shown in Algorithm~\ref{alg:algo}, can be applied on top of any CFX generation algorithm and proceeds as follows. First, an interval abstraction is constructed for the FFNN $\model$ and set $\Delta$; the latter is then checked for soundness (Definition~\ref{def:sound_delta}). Then, the search for a CFX starts. At each iteration, a CFX is generated using the base method and is tested for $\Delta$-robustness using the interval abstraction (Definition~\ref{def:delta_robustness}). If the CFX is robust, then the algorithm terminates and returns the solution. Otherwise, the search continues, allowing for CFXs of increasing \FT{distance} to be found. 
These steps are repeated until a 
\FT{threshold} number of iterations $t$ is reached. As a result, the algorithm is \emph{incomplete}, in that it may report that no $\name$-robust CFX can be found within $t$ steps (while one may exist for larger $t$). 

\begin{algorithm}
\caption{Generation of robust CFXs}\label{alg:algo}
\begin{algorithmic}
\Require FFNN $\model$, 
input $x$ such that $\model(x) = c$,
\State $\qquad\quad$ set of plausible model shifts $\Delta$ and threshold $t$
\State Step 1: build interval abstraction $\abst{\model}{\Delta}$.
\State Step 2: check soundness of $\Delta$
\If{$\Delta$ is sound}
\While{iteration number $< t$}
    \State Step 3: compute CFX $x'$ for $x$ and $\model$
    \State $\quad\quad\quad$using base method
    \If{$\abst{\model}{\Delta}(x') = 1 - c$}
        \State \textbf{return} $x'$
    \Else
        \State Step 4: 
        \FT{increase allowed distance} of next CFX 
        \State Step 5: increase iteration number
    \EndIf
\EndWhile
\EndIf
\State \textbf{return} no robust CFX can be found
\end{algorithmic}
\end{algorithm}

We instantiated Algorithm~\ref{alg:algo} \FT{on the non-robust base methods, i.e., } Wachter et al, Proto and MILP
. We use \textit{Wachter et al-R}, \textit{Proto-R} and \textit{MILP-R}\AR{, respectively,} to denote the resulting algorithms. For each dataset, we use the same $\delta_{max}$ identified in §\ref{ssec:evaluate_rob} to create sound sets of model shifts $\Delta$. The iterative procedure of Algorithm~\ref{alg:algo} generates CFXs of increasing \FT{distance} until the target robustness $\Delta$ is satisfied. To increase the 
\FT{distance} of CFXs for Wachter et al and Proto we iteratively increase the influence of the loss term pertaining to CFX validity. 
For MILP, instead, we require that the probability of the output produced by the classifier to subsequent CFXs increases at each iteration (all test instances are classified as class 0, and the desired class is class 1). More details are included in Appendix~\ref{app:hyperparameters}. \JJ{In practice, the number of iterations will depend on the choice of $\delta$ and the magnitude of step changes of the hyper-parameters, which is specific to each base method (e.g., 25, 6, 35 on average for {Wachter et al-R}, {Proto-R} and {MILP-R}, respectively)}. 

Figures~\ref{fig:analysis_sota}(e-h) show the results obtained. Overall, we can observe that Algorithm~\ref{alg:algo} successfully increases the $\name$-validity of CFXs generated by base methods (compared with Figures~\ref{fig:analysis_sota}(a-d)). MILP\AR{-R} appears to be the best performing algorithm, generating CFXs that always satisfy the given robustness target. The robustness of CFXs computed with Wachter et al and Proto also drastically improves across different datasets. In some cases our algorithm fails to produce robust CFXs \FT{for high values of $\delta$}, yet a considerable improvement in robustness can be observed \FT{overall} (compare, e.g., Figures~\AR{\ref{fig:diab}} and~\ref{fig:diabr}). 
Interestingly, simply by altering the hyperparameters of the \FT{base} methods 
not specifically designed for robustness purposes, they produced more $\name$-robust results than ROAR. 

We also evaluated the extent to which $\Delta$-robustness to smaller model shifts 
can help mitigate the effect of more significant model shifts. To this end, for each base method, we generated $\name$-robust CFXs for a model trained on $\dataset_1$. We then generated a new  model retrained using both $\dataset_1$ and $\dataset_2$ and evaluated the validity of CFXs for the new model. We highlight that this retraining procedure may result in model shifts that are larger than the $\Delta$ targeted 
for the original model (see Appendix~\ref{app:setup} for a detailed analysis). 
 As such, $\name$-robustness may not be guaranteed on the new model. For each algorithm and dataset, we analyse the following metrics: \textbf{vm1}, the percentage of CFXs that are valid on the original model; \textbf{vm2}, the percentage of CFXs that remain valid after retraining; $\ell_1$, the $\ell_1$ distance from the 
 input; \textbf{lof}, the local outlier factor ($+1$ for inliers, $-1$ otherwise), used to test if an instance is within the data manifold. We average $\ell_1$ and \textbf{lof} 
 over the generated CFXs.

Table~\ref{tab:results-NN} reports the results obtained for this second set of experiments. We observe that enforcing $\Delta$-robustness, even for small $\delta$ values, can considerably improve the validity of CFXs in the presence of larger model shifts. Indeed, Algorithm~\ref{alg:algo} increases the number of CFXs that remain valid after retraining 
by 68-100\%. This improvement 
comes at the expense of $\ell_1$ distance, which often increases
. This phenomenon has already been observed in recent work~\cite{Dutta_22}, where robust CFXs for tree classifiers were up to seven times more expensive than the non-robust baselines. The \textbf{lof} score 
tends to remain unchanged 
in many cases. However, for some combinations of 
base methods and datasets, the score drops considerably, suggesting that a better strategy to generate 
CFXs \FT{of increased distance} may exist. Finally, we can observe that our approach often outperforms ROAR, producing CFXs that retain a higher degree of validity after retraining.


\section{Conclusions}
\label{sec:conclusion}
Despite the great deal of attention which CFXs in XAI have received of late,
\FT{SOTA} approaches fall short of providing formal robustness guarantees on the explanations they generate, as we have demonstrated. In this paper 
we proposed $\name$-robustness, a formal notion for assessing the robustness of CFXs with respect to changes in the underlying model. We then introduced an abstraction-based framework to reason about $\name$-robustness and used it to verify the robustness of CFXs and to guide existing methods to find CFXs with robustness guarantees. 

This paper opens several avenues for future work. Firstly, while our experiments only considered FFNNs with ReLU activations
, there seems to be no reason why interval-based analysis for robustness of CFXs could not be applied to a wider range of AI models. 
Secondly, it would be interesting to investigate probabilistic extensions of this work, so as to accommodate scenarios where robustness cannot be always guaranteed. 
Finally, our algorithm for generating $\name$-robust CFXs is incomplete; we plan to investigate whether our abstraction framework can be used to devise complete algorithms with improved guarantees.

\section{Acknowledgements}

Rago and Toni were partially funded by the European Research Council (ERC) under the European Union’s Horizon 2020 research and innovation programme (grant agreement No. 101020934). 
Jiang, Rago and Toni were partially funded by J.P. Morgan and by the Royal Academy of Engineering under the Research Chairs and Senior Research Fellowships scheme. The authors acknowledge financial support from Imperial College London through an Imperial College Research Fellowship grant awarded to Leofante. Any views or opinions expressed herein are solely those of the authors listed.

\bibliographystyle{named}
\bibliography{bib}

\clearpage
\newpage
\appendix

\section*{Appendices}
\appendix

\section{Proofs}
\label{app:proofs}

\textbf{Proof of Lemma~\ref{lemma:bounds}}
\begin{proof}
Combining Definition~\ref{def:set_of_plausible_shifts} with Definition~\ref{def:distance_between_models}, we obtain:

\begin{equation*}
    \left( \sum_{i=1}^{k+1} \sum_{j=1}^{\card{N_i}} \sum_{
    l=1}^{\card{N_{i-1}}} \lvert W_i[j,l] - W'_i[j,l] \rvert^p \right)^{\frac{1}{p}} \leq \delta
\end{equation*}

We raise both sides to the power of $p$:

\begin{equation*}
    \left( \sum_{i=1}^{k+1} \sum_{j=1}^{\card{N_i}} \sum_{
    l=1}^{\card{N_{i-1}}} \lvert W_i[j,l] - W'_i[j,l] \rvert^p \right) \leq \delta^p
\end{equation*}

where the inequality is preserved as both sides are always positive. We now observe this inequation bounds each addend from above, i.e.,

\begin{equation*}
 \lvert W_i[j,l] - W'_i[j,l] \rvert^p \leq \delta^p 
\end{equation*}


Solving the inequation for each addend we obtain $W'_i[j,l] \in [W_i[j,l] - \delta, W_i[j,l] + \delta]$, which gives the result.
\end{proof}

The same result applies to bias values, which were omitted from Definition~\ref{def:distance_between_models} for clarity.\\

\noindent\textbf{Proof of Lemma~\ref{lemma:over-approx}}

\begin{proof}
Lemma~\ref{lemma:bounds} shows that each $\mshift\in\Delta$ maps each weight $W_i[j,l]$ (respectively bias $B_i[j]$) in $\model$ into a closed bounded domain that we denoted as $[\underline{W_i'[j,l]}, \overline{W_i'[j,l]}]$ (respectively $[\underline{B_i'[j]}, \overline{B_i'[j]}]$), for $i \in [k+1]$, $j \in [\card{N_i}]$ and $l \in [\card{N_{i-1}}]$. These domains are used to initialise the corresponding $\mathbf{W}_i[j,l]$ in $\abst{\model}{\Delta}$, which therefore captures all models that can be obtained from $\model$ via $\Delta$ by construction. However, following Remark~\ref{remark:conservative_bounds}, $\abst{\model}{\Delta}$ may also capture additional models for which their $p$-distance from $\model$ is greater than $\delta$.
\end{proof}










\section{Output Range Estimation for INNs}
\label{app:INN}

We use the approach proposed in~\cite{PrabhakarA19} to compute the output reachable intervals for each output of the INN. The output range estimation problem for (ReLU-based) INNs can be encoded in MILP as follows.

The encoding introduces:
\begin{itemize}
    \item a real variable $x_{0,j}$ for $j \in [\card{N_0}]$ used to model the input of the INN;
    \item a real variable $x_{i,j}$ to model the value of each node in $N_i$, for $i \in [k+1]$ and $j \in [\card{N_i}]$,
    \item a binary variable $\delta_{i,j}$ to model the activation state of each node in $N_i$, for $i \in [k]$ and $j \in [\card{N_i}]$.
\end{itemize}

Then, for each $i \in [k]$ and $j \in [\card{N_i}]$ the following set of constraints are asserted:

\begin{equation}
\begin{alignedat}{2}
      & C_{i,j}   &&= \Bigl\{ x_{i,j} \geq 0, x_{i,j} \leq M(1 - \delta_{i,j}), \\
      & && x_{i,j} \leq \sum_{l=1}^{\card{N_{i-1}}} \overline{W_i[j,l]} x_{i-1,j} + \overline{B_i[j]} + M\delta_{i,j}, \\
      & && x_{i,j} \geq \sum_{l=1}^{\card{N_{i-1}}} \underline{W_i[j,l]} x_{i-1,j} + \underline{B_i[j]} \Bigr\}
\end{alignedat}    
\label{eqn:node}
\end{equation}

where $M$ is a sufficiently large constant. Each $C_{i,j}$ uses the standard big-M formulation to encode the ReLU activation~\cite{LomuscioM17} and estimate the lower and upper bounds of nodes in the INN.

Then, constraints pertaining to the output layer $k+1$ are asserted for each class $j \in \card{N_{k+1}}$.

\begin{equation}
\begin{alignedat}{3}
      & C_{k+1,j}   = \Bigl\{ &&x_{k+1,j} &&\leq \sum_{l=1}^{\card{N_{k}}} \overline{W_{k+1}[j,l]} x_{k,j} + \overline{B_{k+1}[j]}, \\
      & && x_{k+1,j} &&\geq \sum_{l=1}^{\card{N_{k}}} \underline{W_{k+1}[j,l]} x_{k,j} + \underline{B_{k+1}[j]}
      \Bigr\}
\end{alignedat}    
\label{eqn:output}
\end{equation}

The output range for a given input $x_0$ and each class $j \in \card{N_{k+1}}$ can be computed by solving two optimisation problems that minimise (respectively maximise) variable $x_{k+1,j}$ subject to constraints~\ref{eqn:node}-\ref{eqn:output}. For more details about the encoding and its properties we refer to the original work~\cite{PrabhakarA19}.

\paragraph{Multiclass classification} The method proposed by~\cite{PrabhakarA19} is directly applicable to the multiclass setting and does not require any change wrt to the original formulation. The semantics of an INN for multiclass problems can be obtained by extending Definition~\ref{def:inn_classification} as follows.

\begin{definition} Consider an input $x\in \mathbb{R}^{\card{N_0}}$, a label $c \in \{1,\ldots,m\}$ and an INN $\intervalnet$. We say that \FL{$\intervalnet$ classifies $x$ as $c$, written $\intervalnet(x) = c$, if ${v}_{k+1,c}^l > {v}_{k+1,j}^u$, for all $j \in \{1,\ldots,m\}$, $j\neq c$.}
\label{def:inn_multiclass_classification}
\end{definition}

The definition of $\name$-robustness transfers to the multiclass semantics straightforwardly; checking whether the robustness property holds simply requires to estimate lower and upper bounds for each output class ($>2$) as needed.

\paragraph{Binary classification with a single output node} As mentioned in the main text, we formalised our results in the context of binary classification implemented via FFNNs with $|N_{k+1}|=2$. However, the most common way to implement such classifiers is to use FFNNs with a single output node with sigmoid activation. This ensures that the output of FFNNs is always in the range $[0,1]$ and allows for a probabilistic interpretation of its value. This is indeed the implementation we used in our experimental analysis.

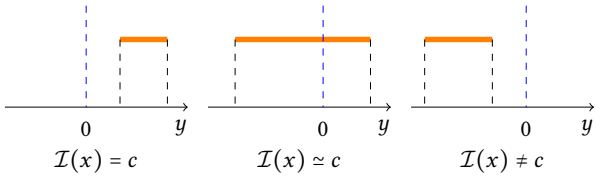
\begin{figure}
    \centering
    \scalebox{0.9}{\begin{tikzpicture}

 
  
 

  

  

\draw[line width=0.8mm, color=orange] (4,0) -- (4.7,0);
\draw[dashed] (4,0) -- (4,-1);
\draw[dashed] (4.7,0) -- (4.7,-1);

\draw[dashed,blue] (3.5,0.5) edge node[above,yshift=-1.3cm]{\color{black}$0$} (3.5,-1);

\draw[->] (2.3,-1) -- (5,-1);
\node[] (phantom_r1) at (4.9,-1.3) {$y$};

\node[] (phantom_1) at (3.65,-1.8) {$\mathcal{I}(x) = c$};


\draw[line width=0.8mm, color=orange] (5.7,0) -- (7.7,0);
\draw[dashed] (5.7,0) -- (5.7,-1);
\draw[dashed] (7.7,0) -- (7.7,-1);

\draw[dashed,blue] (7,0.5) edge node[above,yshift=-1.3cm]{\color{black}$0$} (7,-1);

\draw[->] (5.3,-1) -- (8,-1);
\node[] (phantom_r2) at (7.9,-1.3) {$y$};

\node[] (phantom_1) at (6.65,-1.8) {$\mathcal{I}(x) \simeq c$};


\draw[dashed,blue] (10,0.5) edge node[above,yshift=-1.3cm]{\color{black}$0$} (10,-1);

\draw[line width=0.8mm,color=orange] (8.5,0) -- (9.5,0);
\draw[dashed] (8.5,0) -- (8.5,-1);
\draw[dashed] (9.5,0) -- (9.5,-1);

\draw[->] (8.3,-1) -- (11,-1);
\node[] (phantom_r3) at (10.9,-1.3) {$y$};

\node[] (phantom_1) at (9.65,-1.8) {$\mathcal{I}(x) \neq c$};

\end{tikzpicture}}
    \caption{Graphical representation for the single output node case, assuming that class $c$ corresponds to positive output values. When $\intervalnet(x) = c$, the output range is always greater than zero. Otherwise, we say $\intervalnet(x) \neq c$.
    }
    \label{fig:inn_classfication_sigmoid}
\end{figure}

The interval abstraction can also be applied to this setting, although with two minor modification wrt the formalisation of §\ref{sec:robust_cfx}:
\begin{itemize}
    \item sigmoid activations cannot be directly encoded in the MILP framework used to estimate output ranges of the interval abstraction (cf.~\ref{app:INN}). However, the sigmoid function is invertible over its entire domain; as a result, reasoning about $\name$-robustness can performed on the pre-activation value of the output node without changing its meaning, nor affecting its validity.
    \item As a result, the (pre-activation) output of the interval abstraction is compared to zero instead of the usual $0.5$ threshold determining classification outcome for binary classification using sigmoid (see Figure~\ref{fig:inn_classfication_sigmoid} for a graphical illustration).
\end{itemize}
    
\section{Experimental setup}
\label{app:setup}

\paragraph{Datasets}
For each dataset, we first remove not-a-number values. Then, depending on the actual meanings of the input variables, we categorise them into \emph{continuous}, \emph{ordinal}, \emph{discrete}. For continuous features, we perform min-max scaling; for ordinal features with possible values $[k]$, we encode each value $i$ into an array of shape $(k,)$, with the first $i$ values being 1 and the rest being 0; we one-hot encode the discrete variables. We report the details of each dataset after such preprocessing in Table~\ref{tab:datasets}. For sba dataset, we only use the continuous features. 

\begin{table}[h]
    \centering
    \begin{tabular}{cccc}
        \hline 
        \textbf{dataset} & \textbf{type} & \textbf{instances} & \textbf{variables} \\
        \hline
         diabetes & continuous & 768 & 8 \\
        no2 & continuous & 500 & 7 \\
        sba & continuous & 2102 & 9 \\
        credit & heterogeneous & 1800 & 72 \\\hline
        
    \end{tabular}
    \caption{Dataset details}
    \label{tab:datasets}
\end{table}

\paragraph{Classifiers and training}

We used the sklearn library (https://scikit-learn.org/stable/index.html) for training neural networks with 1 hidden layer. The hyperparameters include number of nodes in the hidden layer, batch size, initial learning rate. The final hyperparameters were found using randomised search and 5-fold cross validation on $\dataset_1$ (we report the classifiers' accuracy and macro-F1 in Table~\ref{tab:clf-5cv}). Accuracy score was used as the model selection criterion. The classifiers with the optimal hyperparameters were then trained on $80\%$ (the training set) of $\dataset_1$, and evaluated on the remaining $20\%$ (the test set). 

\begin{table}[h]
    \centering
    \begin{tabular}{ccc}
        \hline 
        \textbf{Dataset} & \textbf{Accuracy} &  \textbf{Macro-F1}\\
        \hline
        
        diabetes & $0.76\pm0.04$ & $0.73\pm0.04$ \\
        no2 & $0.61\pm0.04$ & $0.61\pm0.04$ \\
        sba & $0.94\pm0.01$ & $0.89\pm0.02$ \\
        credit & $0.75\pm0.02$ & $0.68\pm0.02$ \\ \hline
        
    \end{tabular}
    \caption{Classifier evaluations}
    \label{tab:clf-5cv}
\end{table}

\paragraph{Baseline implementations}
\begin{itemize}
    \item Wachter et al: we use the Alibi library implementation for the method (https://github.com/SeldonIO/alibi). For §\ref{ssec:evaluate_rob}, we change the \emph{target\_proba} hypterparameter to values (0.55 - 0.95) lower than default (1.0) to optimise proximity while maintaining a high validity.
    \item Proto: we use the Alibi library implementation. Hyperparameters of this method are the default values.
    \item MILP: we implemented Definition~\ref{def:cfx} for neural networks, which is also a simplified version of Algorithm 5,~\cite{MohammadiKBV21} satisfying the same \emph{plausibility constraints} in support of heterogeneous datasets.
    \item ROAR: we use their open-sourced implementation: https://github.com/AI4LIFE-GROUP/ROAR. The hyperparameters are set to default values.
\end{itemize}

\paragraph{Evaluation metrics}
Similar to~\cite{Dutta_22}, we use the sklearn implementation of local outlier factor: https://scikit-learn.org/stable/modules/generated/sklearn. neighbors.LocalOutlierFactor.html. Parameters are set to default.

\begin{figure*}[ht]
\centering

  \begin{subfigure}{\columnwidth}
    \includegraphics[width=\linewidth]{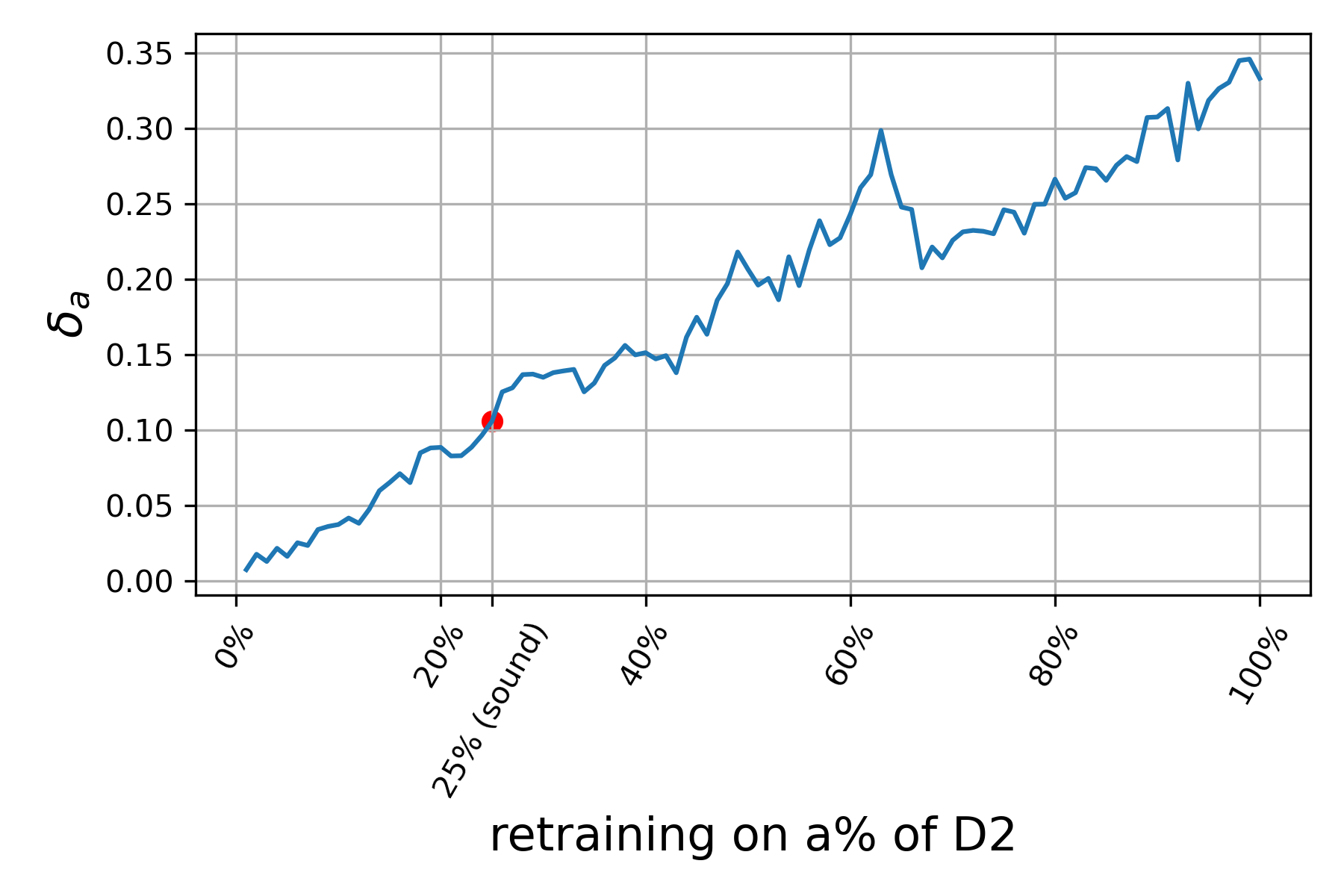}
    \caption{diabetes}
    \label{fig:diab_app}
  \end{subfigure}
  \hfill
  \begin{subfigure}{\columnwidth}
    \includegraphics[width=\linewidth]{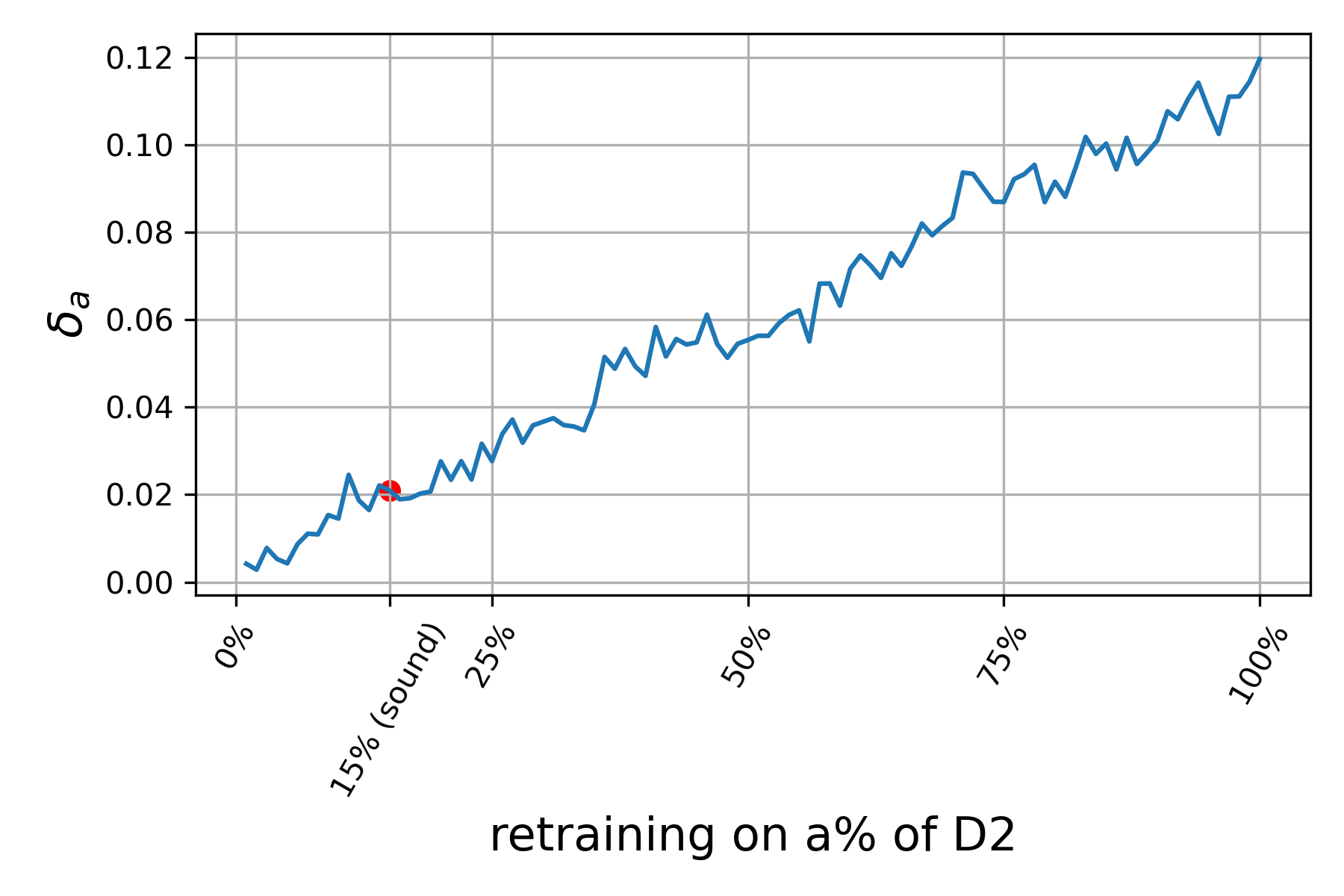}
    \caption{no2}
    \label{fig:no2_app}
  \end{subfigure}
 
  \begin{subfigure}{\columnwidth}
    \includegraphics[width=\linewidth]{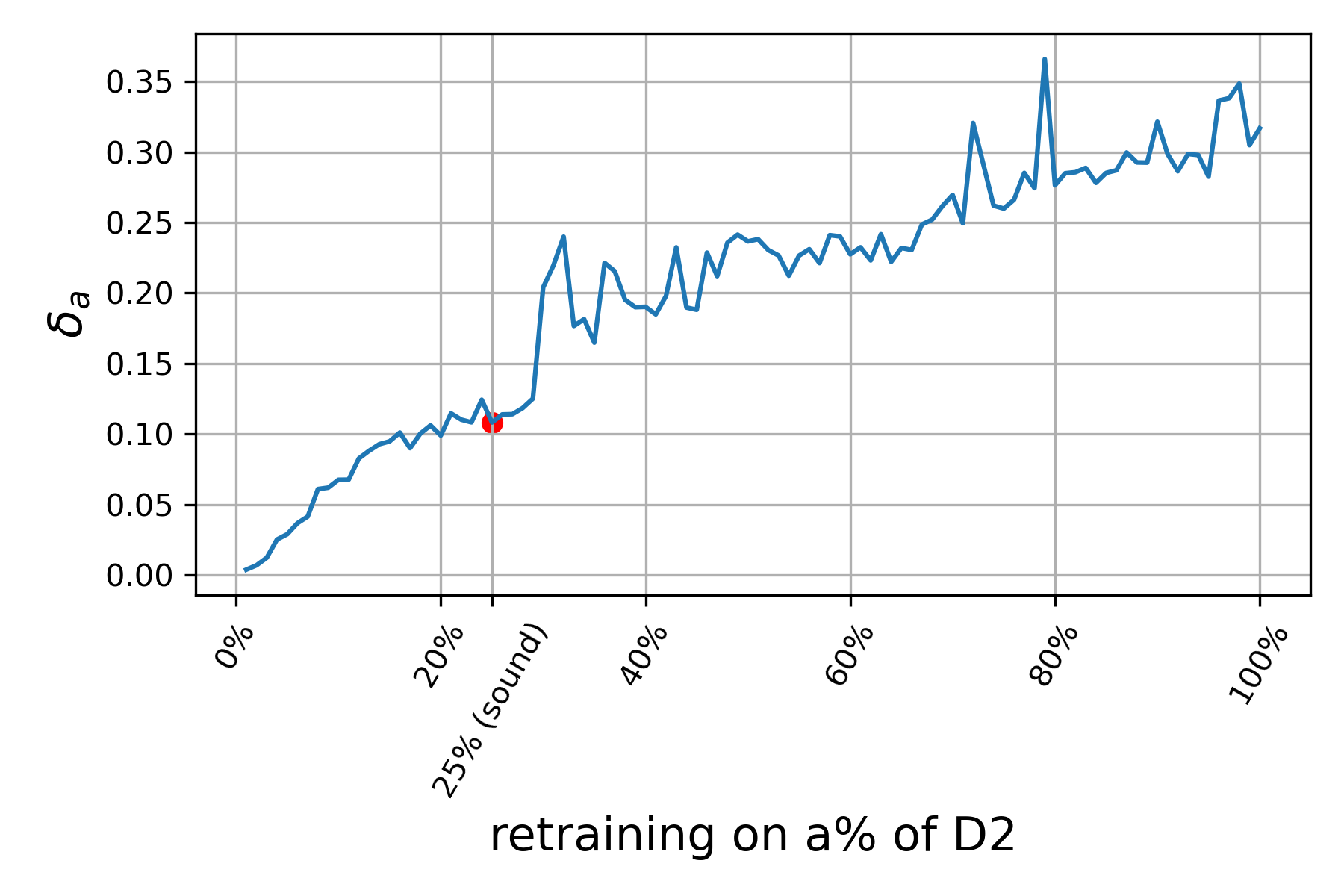}
    \caption{SBA}
    \label{fig:sba_app}
  \end{subfigure}
  \hfill
  \begin{subfigure}{\columnwidth}
    \includegraphics[width=\linewidth]{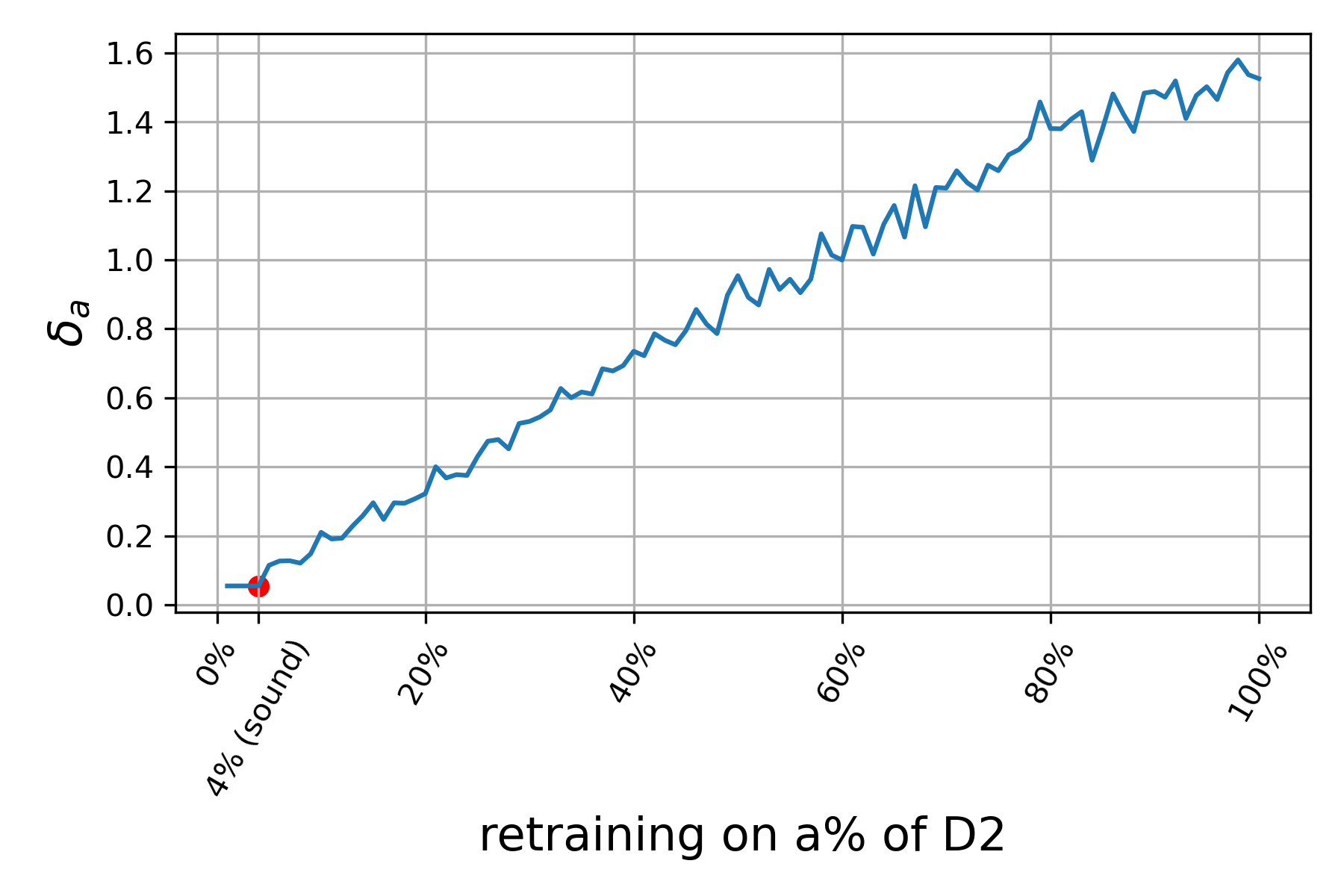}
    \caption{credit}
    \label{fig:cred_app}
  \end{subfigure}
  
  \caption{Plots of $\delta$ values obtained by retraining on increasing portions of $\dataset_2$ described in Step 1, on each dataset. The points highlighted in red in each subfigure corresponds to the $\delta_{max}$ in §\ref{ssec:evaluate_rob} (target $\delta$ in §\ref{ssec:generate_rob}) and the corresponding retraining percentage of $\dataset_2$. $\delta_{max}$ is the largest $\delta$ value that is sound to at least 50 test instances, as also noted in the horizontal axis labels.}
    \label{fig:deltaplot}
\end{figure*}

\paragraph{Retraining and $\delta$ values}
Our retraining procedure follows that of sklearn's \emph{MLPClassifier.partial\_fit()} function. $\delta_{max}$ in §\ref{ssec:evaluate_rob} (also the target $\delta$ in §\ref{ssec:generate_rob}) is obtained by the following procedures. Consider the base model trained on $\dataset_1$, $\model$, and the dataset to retrain on, $\dataset_2$:
\begin{enumerate}
    \item Obtain $\delta_a$ by retraining $\model$ on $a\%$ (initially $a=1$) of $\dataset_2$: randomly select $a\%$ instances from $\dataset_2$ and train on them. Repeat this step for 5 times and get 5 different retrained classifiers, $\model'_i, i \in [5]$. Then, $\delta_a=max(\{\distance{\model}{\model'_i}{p}, i \in [5], p=\infty\})$.
    
    \item Test if model shift $\Delta$ built with $\delta_a$ is sound for at least 50 test instances in $\dataset_1$. 
    
    \item If the condition in step 2 is satisfied, increase $\delta_a$ and repeat step 1 and step 2; if not, $\delta_{max}=\delta_a$.
\end{enumerate}

In order to show how $\delta_a$ (obtained by step 1 above) changes as $a$ increases to $100\%$, we demonstrate the relationship under our settings in Figure~\ref{fig:deltaplot}. It is shown that $\delta_a$ values increase with slight fluctuations as $a$ increases, and the magnitudes depend on the classifier and the dataset. As can be observed, $\delta_{100\%}$ values are always greater than $\delta_{max}$, indicating that the model shifts obtained by retraining on $100\%$ of $\dataset_2$ exceeds those included in $\Delta$ built with $\delta_{max}$, as stated in §\ref{ssec:generate_rob}. Indeed, observing the \textbf{vm2} results of \textbf{MILP-R} from Table~\ref{tab:results-NN} again, we find CFXs that are $100\%$ $\Delta$-robust (as per Figures~\ref{fig:analysis_sota}(e-h)) with a smaller $\delta$ are likely to also be robust to certain instances of $\Delta$ with a larger $\delta$.

\section{Applying $\name$-Robustness as a Filter}
\label{app:DiCE}

\begin{figure}[ht!]
    \centering
    \includegraphics[width=\columnwidth]{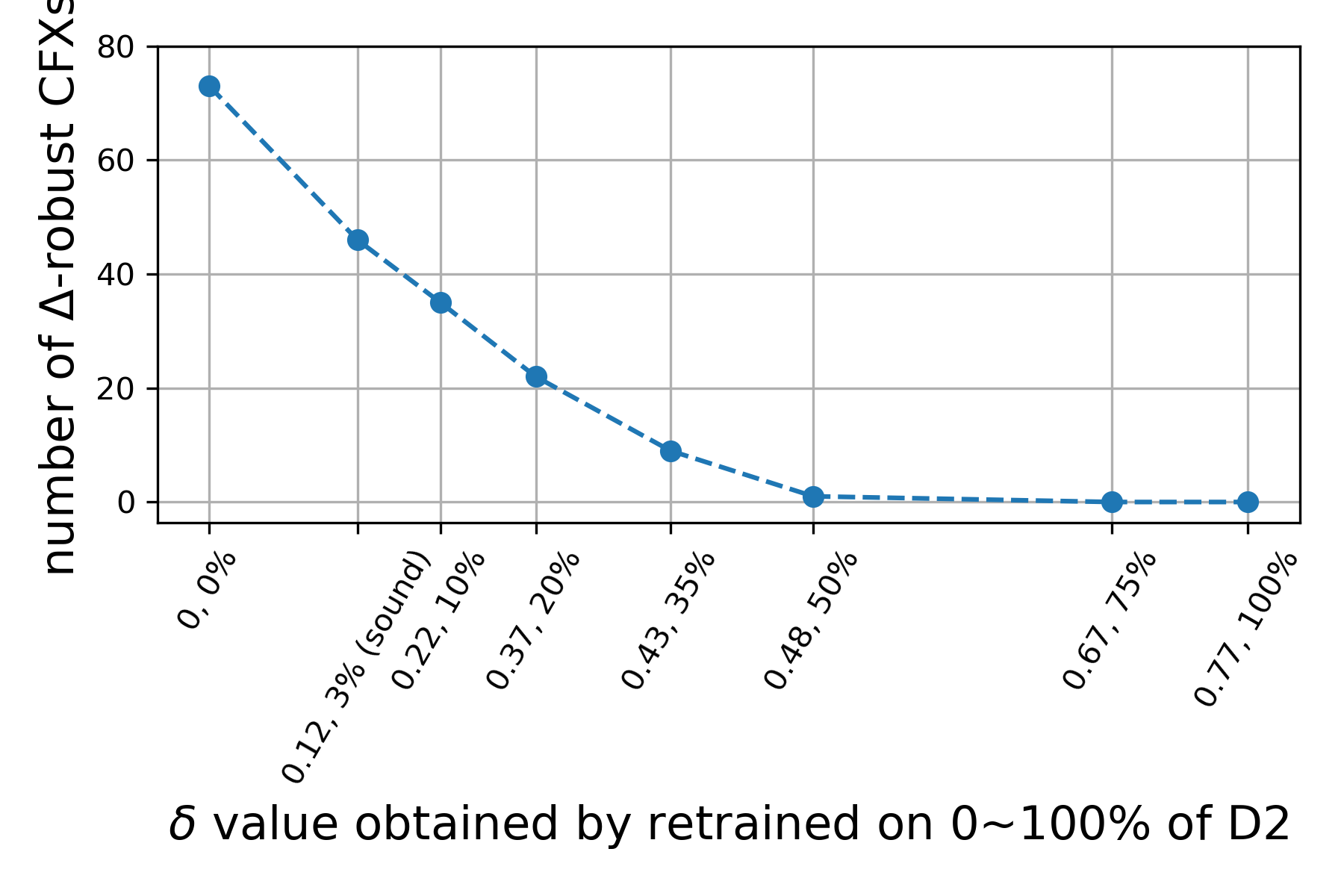}
    \caption{Plots of number of $\Delta$-robust CFXs as $\delta$ increases.}
    \label{fig:dicedemo}
\end{figure}

Similar to evaluating robustness, our approach can also be used as a filter for provably robust CFXs when multiple CFXs are provided for each test instance. To this end, we consider the following setting similar to Appendix~\ref{app:setup}: a neural network is trained on (half of) the HELOC dataset\footnote{FICO Community,\emph{Explainable Machine Learning Challenge}, \url{https://www.kaggle.com/datasets/ajay1735/hmeq-data}, 2019.} to predict whether an applicant can successfully repay the loan given 10 continuous-type credit history information. We consider 5 test instances who failed the loan approval and we generate a diverse set of 20 CFXs for each applicant using DiCE~\cite{mothilal2020explaining}, resulting in a total 100 CFXs. We perform the retraining Step 1 for $a=1~100$, test and report how many CFXs are $\Delta$-robust for some $\delta$ values. We also include the number of valid CFXs on the base model ($\delta=0$). Results are presented in Figure~\ref{fig:dicedemo}. The neural network classifier is trained using PyTorch, the training and retraining settings are implemented in the same way as those of sklearn. We reduced DiCE's hyperparameters \emph{proximity\_weight} to 0.05 and \emph{diversity\_weight} to 1.0 to increase the influence of the prediction correctness loss term and obtain more valid CFXs.

Note that in §\ref{ssec:evaluate_rob} and §\ref{ssec:generate_rob} we only report $\delta$ values whose corresponding $\Delta$ is sound. However, the fact that $\Delta$ is not sound for an instance $x$ means that there exist some model shifts in $\Delta$ under which the classification result of the test instance will change, and in practice there could be only a few such model shifts. Therefore, it could be meaningful to also test such non-sound sets of model shifts, as is the case in this example. We observe that the number of robust CFXs decreases as model changes become larger. Our approach is able to filter out the non-robust CFXs at any given $\Delta$. In reality, as stated in §\ref{sec:intro}, the explanation-providing agent will have an estimate of typical $\delta$ values retrained on certain amount of new data, and the time period to collect such new data. With our approach, the agent could select only the $\Delta$-robust CFXs, and provide a better estimate of how long they are valid for.

\section{Hyperparameter Tuning for Base Methods}
\label{app:hyperparameters}

Concrete implementations of Step 4 in Algorithm~\ref{alg:algo} depend on each base methods' tunable hyperparameters, and are different for Wachter et al.-R, Proto-R, and MILP-R. In this section we introduce detail Step 4 for each of these method. 

{\bf Wachter et al.} In the Alibi library implementation \footnote{https://docs.seldon.io/projects/alibi/en/stable/methods/CF.html}, the loss function of~\cite{Wachter_17} is:
\[
l(x, x') = (\sigma(\model(x')), y_t)^2 + \lambda L1(x, x')
\]
where $\sigma(\model(x'))$ is the value of the neural network's output node and $\sigma$ is the sigmoid function. $\sigma(\model(x')$ can be interpreted as the probability of class 1. $y_t$ is the target probability (of class 1) of the resulting CFX for Wachter et al.-R. In our setting where test instances are of class 0, $y_t$ could take values [0.5, 1.0], the lower $y_t$ is, the easier it is to find a closer (measured by L1 metric) CFX. $\lambda$ is the weight term for proximity. We apply two nested outer loops in search of the optimal $y_t$ and $\lambda$ values to find robust CFXs. For each experiment, we start with the $y_t$ value taken in the non-robust setting, increase $y_t$ by 0.1 until it reaches 1.0. For each $y_t$ to test, we try finding robust CFXs using $\lambda$ values of $\{0.01, 0.05, 0.1, 0.2\}$, where 0.1 is the default value.

{\bf Proto.} In the case of Proto, the loss term pertaining to prediction correctness in the Alibi library implementation is the difference between the output probability of the desired class and the output probability of the undesired class plus $\kappa$ (default 0), a hyperparameter to control the impact of this loss term in the loss function. In Proto-R, we gradually increase $\kappa$ by 0.1 until it reaches 1.0 at each iteration.

{\bf MILP.} For MILP, referring to Definition~\ref{def:cfx}, the $\model(x')=1-c$ condition requires that the lower bound of the output interval be greater than zero which, after the sigmoid output activation, corresponds to a probability greater than or equal to 0.5 for class 1. In this case, this method mostly find CFXs that lie on the decision boundary of the classifier. In MILP-R, to increase the influence of prediction confidence and relax the cost requirement, we require in the MILP encoding that the lower bound of the output interval be greater than $\epsilon$, $\epsilon \geq 0$. At each iteration, we raise $\epsilon$ by 0.2 until it reaches 20.

Concrete values of each of the parameters can be found in the experiments in the accompanying  source codes.


\end{document}